%% file: submission.tex
\documentclass[10pt,twoside]{article}

\usepackage{times}
\usepackage[utf8]{inputenc}
\usepackage[T1]{fontenc}
\usepackage{graphicx}

\usepackage{siunitx}
\usepackage{hyperref}
\usepackage{coria-taln2025}
\usepackage[french]{babel} 
\usepackage{amssymb}

\usepackage{listings}
\usepackage{xcolor}
\usepackage[most]{tcolorbox}

\lstdefinestyle{yaml}{
    backgroundcolor=\color{white},
    basicstyle=\ttfamily,
    keywordstyle=\color{blue},
    stringstyle=\color{red},
    commentstyle=\color{green},
    showstringspaces=false,
}

\lstdefinestyle{json}{
    backgroundcolor=\color{white},
    basicstyle=\ttfamily,
    keywordstyle=\color{blue},
    stringstyle=\color{red},
    commentstyle=\color{green},
    showstringspaces=false,
}

\lstdefinestyle{python}{
    backgroundcolor=\color{white},
    basicstyle=\footnotesize\ttfamily,
    keywordstyle=\color{blue},
    stringstyle=\color{red},
    commentstyle=\color{green},
    showstringspaces=false,
}

\newtcolorbox{pydanticbox}[2][]{%
  colback      = blue!5!white,
  colframe     = blue!75!black,
  fonttitle    = \bfseries,
  colbacktitle = blue!85!black,
  title        = #2,#1,
  fontupper    =\ttfamily\fontsize{8pt}{8pt}\selectfont
}

\newtcolorbox{promptwheelbox}[2][]{%
  colback      = yellow!5!white,
  colframe     = yellow!75!black,
  fonttitle    = \bfseries,
  colbacktitle = yellow!85!black,
  title        = #2,#1,
  fontupper    =\ttfamily\fontsize{8pt}{8pt}\selectfont
}

\newtcolorbox{smallpromptwheelbox}[2][]{%
  colback      = yellow!5!white,
  colframe     = yellow!75!black,
  fonttitle    = \bfseries,
  colbacktitle = yellow!85!black,
  title        = #2,#1,
  fontupper    =\ttfamily\fontsize{9pt}{10.2pt}\selectfont
}
\newtcolorbox{promptbox}[2][]{%
  colback      = green!5!white,
  colframe     = green!75!black,
  fonttitle    = \bfseries,
  colbacktitle = green!85!black,
  title        = #2,#1,
  fontupper    =\ttfamily\fontsize{8pt}{8pt}\selectfont
}
\usepackage{caption}
\usepackage{subcaption}
\usepackage{float}
\restylefloat{table}
\newfloat{code}{htbp}{lop}
\floatname{code}{Algorithme}

\let\oldtexttt\texttt
\renewcommand{\texttt}[1]{{\small\oldtexttt{#1}}}

\newcommand{\cpt}[0]{\textsc{PPE}}
\newcommand{\ift}[0]{\textsc{ASI}}

\newcommand{\mistralsept}[0]{\texttt{\small{Mistral-7B-Instruct-v0.3}}}
\newcommand{\closedruna}[0]{\texttt{O\_FT-Fermé-run1}}
\newcommand{\closedrunb}[0]{\texttt{O\_FT-Fermé-run2}}
\newcommand{\closedrunc}[0]{\texttt{O\_FT-Fermé-run3}}
\newcommand{\openruna}[0]{\texttt{O\_FT-Ouvert-run1}}
\newcommand{\openrunb}[0]{\texttt{O\_FT-Ouvert-run2}}
\newcommand{\openrunc}[0]{\texttt{O\_FT-Ouvert-run3}}
\newcommand{\doc}[1]{\texttt{\small #1}}
\newcommand{\format}[1]{\texttt{\footnotesize #1}}
\newcommand{\rang}[1]{$_{ (#1)}$}

\usepackage{cleveref}

\title{O\_FT@EvalLLM2025 : étude comparative de choix de données et de stratégies d'apprentissage pour l'adaptation de modèles de langue à un domaine}

\author{Ismaël Rousseau\up{1{\scriptsize a}}\quad Claire Perroux\up{1{\scriptsize b}}\quad Pierre Adam\up{1{\scriptsize a}}\quad \\ Thomas Girault\up{2}\quad Lionel Delphin-Poulat\up{1{\scriptsize a}}\quad Morgan Veyret\up{1{\scriptsize a}}\quad \\ Gwénolé Lecorvé\up{1{\scriptsize a}}\quad Géraldine Damnati\up{1{\scriptsize a}}\\
  {\small
    (1) Orange Research, \up{\up{\tiny a}}Lannion / \up{\up{\tiny b}}Châtillon, France \\ 
    (2) Ouest-France, Rennes, France \\ 
    \texttt{
      prenom.nom@orange.com, prenom.nom@ouest-france.fr \\ 
}}}

\begin{document}
\maketitle

\resume{
  Ce document présente les travaux réalisés par l'équipe O\_FT conjointe à Orange et Ouest-France sur l'adaptation de modèles de langue au domaine de la défense dans le cadre du challenge EvalLLM2025. Ces travaux se sont concentrés sur l'adaptation du modèle \texttt{Mistral-7B-Instruct-v0.3} avec des techniques classiques de poursuite du pré-entraînement et d'affinage sur instructions. L'essentiel de nos travaux a porté sur la constitution, génération et sélection de données pour ces deux étapes ainsi que pour l'évaluation des modèles. Les expériences montrent que nos modèles adaptés ont de meilleures de connaissances de fond et une meilleure capacité de traitement de tâches sur le domaine de la défense, ainsi que des performances comparables (voire supérieures) sur des connaissances ou capacités généralistes. Mis au regard des empreintes carbones de nos adaptations, ces travaux démontrent ainsi la viabilité de l'adaptation à un domaine de modèles relativement petits.
}

\abstract{O\_FT@EvalLLM2025: Comparative Study of Data for Continued Pre-Training and Instruction-Tuning for Language Model Domain Adaptation.}{
This paper presents the work carried out by the O\_FT team, joint with Orange and Ouest-France, on adapting language models to the defense domain as part of the EvalLLM2025 challenge. This work focused on adapting the \texttt{Mistral-7B-Instruct-v0.3} model using classical techniques of continued pre-training and instruction-tuning. The core of our efforts is based on collecting, generating, and selecting data for these two stages as well as for model evaluation. Experiments show that our adapted models have better domain-specific knowledge and improved domain-specific task processing skills, along with comparable (or even superior) performance on general knowledge and skills. Considering the carbon footprint of our adaptations, this work demonstrates the feasibility of domain adaptation for relatively small models.
}

\motsClefs
  {Affinage de modèle de langage, Adaptation au domaine, Génération d'instructions}
  {Language model fine-tuning, Domain Adaptation, Instruction Generation}

\section{Introduction}

La participation d'Orange au challenge EvalLLM2025 s'est concentrée sur l'adaptation du modèle \texttt{Mistral-7B-Instruct-v0.3}, et ce dans les deux configurations proposées par le challenge, à savoir l'une restreinte aux seules données fournies par les organisateurs et l'autre ouverte, où nous avons exploité des données complémentaires.
Sur la base de précédents travaux d'adaptation au domaine des télécommunications~\cite{barboule2024telcolm,shahid2025large}, l'adaptation a consisté à enchaîner une étape de Poursuite du Pré-Entraînement (\cpt, \textit{continued pre-training} en anglais) et une étape d'Affinage sur Instructions (\ift, \textit{instruction  fine-tuning} en anglais). Les deux étapes ont été réalisées en affinant tous les poids (\textit{full fine-tuning}), par opposition aux approches d'adaptation à faible rang.

Dans le cadre de la participation à ce challenge, nous avons conclu un accord de collaboration entre les équipes de recherche d'Orange Innovation et de Ouest-France. Ouest-France nous a donné accès à un certain nombre de contenus et a contribué également à la préparation des données et des instructions. La préparation des données fournies par l'AMIAD, la génération d'instructions et les apprentissages et évaluations ont été conduits à Orange.

Les choix de soumission de modèles ont principalement été guidés par la possibilité d'analyser l'impact de la phase de  poursuite du pré-entraînement d'une part et l'impact de la diversité des instructions pour l'affinage sur instructions d'autre part. En effet, la particularité de l'adaptation à un domaine est que l'on cherche à la fois à faire acquérir des connaissances au modèle tout en maintenant ses capacités à répondre correctement à des instructions diverses. Nous avons apporté un soin particulier à la mise en place d'évaluations variées pour tenter de mesurer les connaissances acquises par les modèles adaptés.

Nous présentons dans ce rapport les étapes de préparation des données à la section~\ref{sec:data}, puis à la section~\ref{sec:instruct} la méthodologie employée pour générer des instructions pour l'adaptation. Les détails concernant les processus de fine-tuning sont fournis à la section~\ref{sec:adapt} et enfin, nous communiquons les évaluations en termes de performances et de consommation énergétique à la section~\ref{sec:eval}.

\input{1_data}

\input{2_instructions}
\input{3_finetuning}
\input{4_eval}
\section{Discussion}
Plusieurs points dans le processus complet peuvent encore être améliorés. Par exemple le nettoyage et le filtrage des documents en amont est perfectible (nous aurions pu par exemple concaténer les segments courts plutôt que de les filtrer). Il serait intéressant par ailleurs de mieux exploiter les images et les tableaux. 
Comme présenté en section \ref{sec:pretraitement_donnees}, le texte des fichiers PDF est extrait directement ou à l'aide d'un OCR. Cependant, nous n'avons pas réalisé de traitements supplémentaires pour prendre correctement en compte les tableaux. De même, nous n'avons pas du tout exploité les images. Il serait également possible de générer des descriptions textuelles des images afin de venir enrichir notre jeu de données avec des informations supplémentaires.
Au niveau de la quantité d'instructions générées, nous générons actuellement une seule instruction par segment. Or, les segments contiennent dans la majorité des cas assez de contenu pour générer plusieurs instructions portant sur des connaissances différentes. 
En ce qui concerne l'adaptation en elle même, nous avons concentré nos efforts sur les connaissances et le respect des instructions, mais il serait également intéressant de travailler sur le style des réponses fournies par les modèles adaptés (avec notamment des approches par renforcement). Actuellement, nos modèles adaptés produisent des réponses plutôt brèves en comparaison avec le modèle \mistralsept~original. Lors d'une évaluation humaine des modèles de type Arène (non incluse dans ce rapport) par des collaborateurs extérieurs au challenge, il s'agissait d'une remarque récurrente venant pénaliser les modèles produits. De plus, le formatage des réponses en MarkDown de certains modèles comme \texttt{GPT-4.1-mini} a été apprécié et a pu faire basculer la préférence des utilisateurs pour ce modèle quand les deux réponses étaient factuellement correctes.
Enfin, nous nous sommes concentrés sur des approches \textit{Full Fine Tuning} et il serait intéressant de confronter nos conclusions avec des approches plus efficientes de type LoRa afin de réduire l'empreinte carbone du processus.
\section{Conclusion}
Nous avons présenté  le détail des approches mises en œuvre pour l'adaptation du modèle \mistralsept~au domaine de la Défense, en mode \texttt{Fermé} où seules les données fournies par les organisateurs du challenge EvalLLM2025 ont été utilisées, puis en mode \texttt{Ouvert} avec des données additionnelles.  L'effort a porté sur la constitution d'un ensemble d'instructions diversifiées pour l'apprentissage, ainsi que sur la mise en place d'un benchmark couvrant plusieurs tâches et compétences pour évaluer nos modèles. Les expériences montrent que nos modèles adaptés ont de meilleures de connaissances de fond et une meilleure capacité de traitement de tâches sur le domaine
de la défense, ainsi que des performances comparables (voire supérieures) sur des connaissances
ou capacités généralistes.



\bibliographystyle{coria-taln2025}
\bibliography{submission}
\input{5_annexes}
\end{document}

%% file: 1_data.tex
\section{Préparation des données}
\label{sec:data}

Cette section décrit le processus suivi pour obtenir des données textuelles exploitables pour nos expériences, que ce soit à la poursuite du pré-entraînement ou à la génération d'instructions (sujet abordé dans la section qui suivra). Après avoir énuméré les sources de données considérées, nous détaillons leur nettoyage, puis leur partitionnement.

\subsection{Description de la base documentaire}

\paragraph{Documents fournis par les organisateurs}

Les documents fournis par les organisateurs proviennent du site internet du ministère des armées (3257 documents \texttt{minarm}), de sources internes (2864 documents \texttt{interne}),  ainsi que du portail "Armée" de Wikipedia (2847 documents). La nature et le contenu des documents sont assez variés et une description plus détaillée par source est fournie dans la table~\ref{tab:raw_documents}. Ces documents se présentent sous format PDF, Word, MarkDown, texte et PowerPoint. Les statistiques par format sont données dans la table~\ref{tab:raw_documents_types_stats}. 

\begin{table}[h!]
    \centering
    \begin{tabular}{l|rcp{5.5cm}}
         \multicolumn{1}{c|}{\textbf{Jeu de données}} & \textbf{Nb doc} & \textbf{Formats} & \textbf{Description} \\
         \hline
         \doc{minarm\_AdT} & 500 & \format{md} & Organisation, matériel et actualités de l'armée de terre \\
         \doc{minarm\_air} & 499 & \format{md} & Actualités et infos sur l'armée de l'air \\
         \doc{minarm\_DGA} & 509  & \format{md} & Organisation et programmes de la Direction Générale de l'Armement \\
         \doc{minarm\_EMA} & 245 & \format{md} & Activités et actualités de l'État-Major des Armées \\
         \doc{minarm\_marine} & 500 & \format{md} & Organisation, équipements et missions de la Marine Nationale \\
         \doc{minarm\_other1} & 495 & \format{md} & Informations diverses sur les entités du Ministère des Armées \\
         \doc{minarm\_SGA} & 509 & \format{md} & Activités du Secrétariat Général pour l'Administration \\
         \hline
         \doc{interne\_divers} & 17 & \format{pdf} & Notes et rapports divers \\
         \doc{interne\_EMA} & 237 & \format{pdf/docx} & Doctrine militaire \& RETEX \\
         \doc{interne\_formation} & 57 & \format{pdf/docx/pptx} & Documents de formation au numérique \\
         \doc{interne\_magazines} & 38 & \format{pdf} & Magazines Air actualités, Esprit Défense et SOUTENIR \\
         \doc{interne\_NP\_BOC} & 2\,483 & \format{pdf} & Bulletin officiel des armées (arrêtés, décisions, instructions, circulaires etc.) \\
         \doc{interne\_reglements} & 32 & \format{pdf} & Règlements divers \\
         \hline
         \doc{wp\_portail\_armée} & 2\,847 & \format{txt} & Pages Wikipédia du portail "Armée"
    \end{tabular}
    \caption{Présentation des documents bruts fournis par les organisateurs}
    \label{tab:raw_documents}
\end{table}


\begin{table}[h!]
    \centering
\begin{tabular}{rccccc}
\textbf{Format} & MarkDown & PDF  & Word  & PowerPoint  & Texte \\
       & \format{md} & \format{pdf} & \format{docx} & \format{pptx} & \format{txt} \\
\hline
\textbf{Nb documents bruts} & 3\,235 & 2\,789 & 31 & 2 & 2843 \\
\end{tabular}

    \caption{Nombre de documents bruts par type de fichier}
    \label{tab:raw_documents_types_stats}
\end{table}

\paragraph{Documents additionnels fournis par Ouest-France}
Des données textuelles additionnelles ont été collectées au sein des archives de presse du groupe SIPA-Ouest-France. Le but est de compléter la terminologie et les connaissances fondamentales du domaine, tout en intégrant une perspective journalistique sur les faits d'actualité récents.
Ce corpus agrège un total de 10\,910 articles issus du journal Ouest-France (7\,800 articles), du blog "Ligne de défense" (755 articles) et du magazine "Le Marin" (2\,355 articles). Le blog "Ligne de défense"\footnote{\url{https://lignesdedefense.ouest-france.fr/}} regroupe des articles rédigés par Phillipe Chapleau, journaliste spécialisé dans les questions de politique étrangère et de défense. 

Une sélection a été faite pour retenir les articles publiés entre le 1er janvier 2022 et le 16 mai 2025. Le choix de ces dates a été défini en faisant l'hypothèse que les données plus anciennes ont probablement été vues par modèle à adapter lors de son entraînement. 

Pour obtenir les données du journal Ouest-France, une sélection thématique a été effectuée à l'aide d'un classifieur fondé sur un modèle CamemBERT~\cite{Martin_2020} entraîné pour identifier les articles correspondant aux catégories \textit{International Press Telecommunications Council} (IPTC)~\cite{iptc2020newscodes} suivantes : 
\begin{itemize}
    \item \doc{/Guerre et conflits/Armement},
    \item \doc{/Politique/Défense},
    \item \doc{/Économie/Industrie de défense}.
\end{itemize} 
Enfin, un filtrage supplémentaire a été réalisé sur ces mêmes données pour ne conserver que les articles avec une longueur variant entre 200 et 3 000 mots.




\subsection{Pré-traitement des données}
\label{sec:pretraitement_donnees}
\subsubsection{AMIAD}
Les fichiers PDF contenant du texte pouvant directement être extrait ont été traités à l'aide de PyMuPDF\footnote{\url{https://github.com/pymupdf/PyMuPDF}}. Les fichiers PDF dont les textes ne sont pas directement accessibles sont traités par la solution d'OCR DocTR \footnote{\url{https://github.com/mindee/doctr}}. Nous regroupons les textes qui sont proches géométriquement en blocs de texte. Ces blocs de texte sont ensuite concaténés, avec un retour à la ligne comme séparateur. Les images (photographies, cartes, etc.) n'ont pas été traitées en tant que telles mais le texte inclus dans les images a pu être traité par l'OCR.

L'extraction du texte pour les fichiers Word a été réalisée via la librairie python \textit{docx2txt}\footnote{\url{https://github.com/ankushshah89/python-docx2txt}}. Les fichiers PowerPoint n'ont pas été utilisés.

Les documents ont ensuite été divisés en segments plus petits pour créer des unités textuelles ciblées facilitant la génération d'instructions, tout en respectant les contraintes de taille de contexte imposées par le modèle \texttt{Mistral-7B-Instruct-v0.3}.

Ainsi, nous avons décidé de couper :
\begin{itemize}
    \item \format{.pdf} et \format{.docx} : par page
    \item \format{.md} : par blocs de 8192 tokens maximum, en coupant à une fin de section
    \item \format{.txt} : par blocs de 8192 tokens maximum, en coupant à un saut de ligne.
\end{itemize}

\vspace{0.5cm}
Après cette découpe, certains segments étaient trop courts et ne contenaient plus d'information substantielle, nous avons donc retiré du corpus les segments de moins de 350 caractères. De plus, nous avons essayé au mieux de retirer les entêtes sur les documents web du ministère de l'armée. 
Certaines n'ont pas pu être nettoyées avec de simples expressions régulières et ont donc été conservées dans les documents, ce qui a pu ensuite poser problème lors de la génération des instructions. 

La table \ref{tab:chunks_stats} regroupe les statistiques sur les segments produits en fonction des différentes sources fournies par les organisateurs.

\begin{table}[h!]
    \centering
    \begin{tabular}{l|r|r@{~}r|r|r}
         \multicolumn{1}{c|}{\textbf{Jeu de données}} & \textbf{Nb seg}~~ & \multicolumn{2}{c|}{\textbf{Seg. écartés}} & \multicolumn{1}{p{1.2cm}|}{\textbf{Lg. moy. par seg. (tokens)}} & \multicolumn{1}{p{1cm}}{\centering \textbf{Seg. par doc.}} \\
         \hline
         \doc{minarm\_AdT} & 531 & 0 & (0.0\%) & 2\,874 & 1.1 \\
         \doc{minarm\_air} & 517 & 1 & (0.2\%) & 2\,566 & 1.0 \\
         \doc{minarm\_DGA} & 517 & 3 & (0.6\%) & 1\,873 & 1.0 \\
         \doc{minarm\_EMA} & 258 & 0 & (0.0\%) & 3\,149 & 1.1 \\
         \doc{minarm\_marine} & 517 & 0 & (0.0\%) & 2\,067 & 1.0 \\
         \doc{minarm\_other1} & 728 & 2 & (0.3\%) & 3\,228 & 1.5 \\
         \doc{minarm\_SGA} & 564 & 0 & (0.0\%) & 2\,527 & 1.1 \\
         \hline
         \doc{interne\_divers} & 630 & 52  & (8.3\%) & 767 & 41.3 \\
         \doc{interne\_EMA} & 14\,579 & 2\,487  & (17.1\%) & 772 & 51.9 \\
         \doc{interne\_formation} & 165 & 63 & (38.2\%) & 698 & 1.9 \\
         \doc{interne\_magazines} & 1\,867 & 184 & (9.9\%) & 815 & 52.6 \\
         \doc{interne\_NP\_BOC} & 28\,064 & 4\,795 & (17.1\%) & 810 & 9.4 \\
         \doc{interne\_reglements} & 2\,953 & 119 & (4.0\%) & 838 & 97.7 \\
         \hline
         \doc{wikipedia\_portail\_armée} & 2975 & 44 & (1.5\%) & 2\,228 & 1.0
    \end{tabular}
    \caption{Statistiques sur les segments}
    \label{tab:chunks_stats}
\end{table}

\subsubsection{Cas particuliers}
Certains documents ont été identifiés comme contenant plus d'informations essentielles que d'autres et ont donc été traités avec plus d'attention. C'est notamment le cas de plusieurs fichiers contenant des listes d'acronymes ou de traductions, à savoir en détails~:
\begin{itemize}
    \item Acronymes : \doc{\small 20240419\_NP\_Livret\_accueil\_DGA\_MI.pdf}, \doc{Chiffres clés de la Défense - 2024 FR.pdf}, \doc{Manuel de droit des opérations militaires.pdf}, \doc{Répertoire acronymes FR-ONU-OTAN\_version 2020.pdf}
    \item Traduction : \doc{20231220\_Vocabulaire militaire français-anglais.pdf}
\end{itemize}

Dans ces deux cas, les fichiers sont composés de nombreux tableaux ou autres formats de correspondance entre les acronymes et leur signification ou bien entre la traduction anglaise et française. Nous avons utilisé \texttt{GPT-4o-mini} afin d'extraire proprement chaque acronyme et traduction. Pour les acronymes, nous gardons en mémoire toutes les significations différentes rencontrées.

\subsubsection{Articles Ouest-France}

Pour cette source de données, les articles ont été conservés dans leur intégralité. Ils sont stockés dans une base de données selon un format spécifiquement conçu pour l'archivage des contenus éditoriaux. Un pré-traitement a été réalisé de façon à nettoyer les balises HTML et les éventuelles métadonnées résiduelles dans les données extraites. Le corpus final est ainsi constitué de 10\,910 segments. Une sélection préliminaire avait déjà été réalisée en amont, retenant uniquement les articles dont la longueur se situait entre 200 et 3 000 mots.

\subsection{Statistiques sur les données brutes après pré-traitements}
Au total, le corpus brut est constitué de 54\,865 segments de texte issus des données fournies par les organisateurs et de 10\,910 segments issus des données journalistiques collectées par Ouest-France.

\begin{table}[h!]
    \centering
    \begin{tabular}{c|ccc}
          & \textbf{Gen AMIAD} & \textbf{Gen OF}  \\
         \hline
         Nb segments & 54\,865 &  10\,910 \\
         Nb mots & 18,7M & 5,9M \\
         Nb. tokens & 48,6M & 11,5M \\
    \end{tabular}
    \caption{Statistiques sur les données brutes pré-traitées pour les deux sources de données}
    \label{tab:instruct-tasks}
\end{table}

Pour chaque jeu de donnée, les segments sont divisés en une partition d'entraînement (80\%), de validation (10\%) et de test (10\%). Cette division est effectuée au niveau des segments et non au niveau des documents ; il est ainsi possible pour un même document d'avoir des segments présents dans différentes partitions différentes, ce qui pourrait être revu dans une version  ultérieure.

%% file: 2_instructions.tex
\section{Génération des instructions}
\label{sec:instruct}
Le corpus d'instructions que nous avons mis en place a été constitué avec quatre approches principales :
\begin{itemize}
    \item Génération d'instructions synthétiques avec un modèle génératif (LLM)
    \item Constructions d'instructions à partir de patrons sur les fichiers acronymes et traduction
    \item Construction d'instructions "longues"
    \item Génération d'instructions spécifiques aux données Ouest-France
\end{itemize}
Les sections suivantes décrivent la méthodologie de chacune des approches.

\subsection{Instructions synthétiques générées par LLMs}
\label{sec:instructgpt}
Nous définissons ici une instruction comme un triplet (amorce système, question utilisateur, réponse). La section \ref{sec:instructcuration} décrit l'utilisation de \texttt{GPT-4.1-mini} afin de générer la question utilisateur ainsi que la réponse de référence, en nous basant sur les segments obtenus après le traitement des documents décrits en section \ref{sec:pretraitement_donnees} pour quatre tâches. La sections \ref{sec:instructof} décrit l'utilisation de \texttt{GPT-4o-mini} pour la génération d'une cinquième tâche propre aux données journalistiques. Le processus de génération des amorces système utilise une autre méthode de génération détaillée en section \ref{sec:amorces_systemes}.

\subsubsection{Définition des tâches et curation des instructions générées}
\label{sec:instructcuration}
Nous avons utilisé le modèle \texttt{GPT-4.1-mini}\footnote{La version exacte utilisée est : \texttt{gpt-4.1-mini-2025-04-14}} d'OpenAI afin de générer des instructions sur quatre tâches cibles: le \textbf{résumé} de texte, le \textbf{titrage}, les \textbf{questions à choix multiples} (QCM), la réponse à des \textbf{questions factuelles}.

Pour chaque segment, nous utilisons \texttt{GPT-4.1-mini} afin de générer une question utilisateur ainsi que la réponse de référence.
La constitution des prompts pour cette étape de génération a été guidée par deux objectifs :
\begin{enumerate}
    \item garantir une diversité dans les formulations des instructions générées pour une même tâche
    \item maîtriser le format des instructions produites pour faciliter l'étape de curation et les exploiter plus facilement dans une optique d'évaluation
\end{enumerate}

\paragraph{Diversité des formulations :} Le prompt utilisé pour la génération des instructions est donné dans le cadre suivant. Le champ \texttt{\textbf{additional information}} est spécifique à chaque tâche cible et a été conçu pour favoriser la diversité des instructions générées. Il contient des directives précises sur la façon de formuler les questions et réponses selon le type de tâche. Pour cela, nous avons défini des grammaires pour produire cette partie du prompt système. Par exemple, pour le cas du résumé ces informations additionnelles permettent d'obtenir des formulations très concises comme "\textsl{Produire un résumé.}" ou des formulations plus détaillées comme \og{}\textsl{Pourriez-vous fournir un résumé de ce texte concernant la reconnaissance des harkis et les implications politiques et morales de leurs décisions durant et après la guerre d'Algérie ?}\fg{}. Les grammaires sont détaillées en annexe \ref{annex:grammar_additional_information} pour chacune des tâches cibles.

\begin{promptbox}[colback=white]{Prompt système pour la génération d'instructions}
You are \textbf{\{persona\}}, with an exhaustive knowledge of LLM instructions generation, your role is to create a set of high-quality \textbf{\{instruction type\}} instructions on the text input provided by the user. All your outputs should be written in french. \textbf{\{additional information\}}

\textbf{Persona} : a senior expert specializing in the defense industry and military affairs. Your expertise covers weapon systems (land, naval, air), emerging technologies (AI, robotics, cyber, space), industrial processes, military R\&D, operational doctrines and strategies, intelligence, as well as the economic aspects of the sector (markets, actors, regulations). You have a thorough understanding of geopolitical issues, international relations, and defense policies
\end{promptbox}

\paragraph{Format des instructions produites}
Afin de maîtriser le format des instructions générées, nous avons exploité le mode de génération structurée d'OpenAI\footnote{\url{https://platform.openai.com/docs/guides/structured-outputs}}. Ce mode de génération prend en entrée une amorce système, une amorce utilisateur ainsi qu'une classe \texttt{BaseModel} \textit{Pydantic}\footnote{\url{https://docs.pydantic.dev/latest/}} contenant le nom des différents champs attendus lors de la génération, ainsi que leur type et leur description. Obtenir une sortie structurée permet à la fois d'exploiter les résultats de génération plus facilement par la suite pour les adapter au \textit{chat template}, et également de les parser pour les retravailler si besoin. Par exemple, pour le cas des questions factuelles, on spécifie un format attendu dans lequel apparaissent distinctement le fait retenu, son type (date, lieu, etc..)  et la question posée. Le détail complet des classes \textit{Pydantic} avec la spécification des champs attendus est fourni en annexe \ref{annex:pydantic_classes}.

 \paragraph{Curation des instructions générées}
Un problème récurrent lors de la génération d'instructions, même avec des efforts sur la formulation de l'amorce, est la tendance de certaines instructions à référencer explicitement le document plutôt que de l'exploiter comme ressource contextuelle. 
A contrario, il arrive fréquemment que l'instruction générée fasse référence à un document alors que l'on souhaite produire une instruction sans contexte. Nous filtrons donc toutes les instructions qui contiennent les expressions suivantes pour les tâches de \textit{QCM} et de \textit{questions factuelles} qui doivent être réalisés sans contexte additionnel : \og{}mentionne\fg{} ;
\og{}selon le\fg{} ;
\og{}d'après le\fg{} ;
\og{}le document\fg{} ;
\og{}ce document\fg{} ;
\og{}le texte\fg{} ;
ou \og{}ce texte\fg{}.

D'autres cas plus complexes comme le cas suivant où la question contient une coréférence demeurent complexes à filtrer et n'ont pas été écartés dans cette étude.
Exemple de QCM contenant une mention de coréférence ("le message") pour lequel il est impossible de répondre sans contexte : 
\begin{verse}
    \og{}\textsl{Que signifie le message survolé par l'avion du général Leclerc ?\newline a - Un appel à la résistance des Parisiens.\newline b - Une annonce d'arrivée imminente des Alliés.\newline c - Un ordre de retrait pour les troupes françaises.\newline d - Une déclaration de guerre.\newline e - Une invitation à abandonner la lutte.}\fg{}
\end{verse}

%


\subsubsection{Instructions spécifiques générées à partir des données Ouest-France}
\label{sec:instructof}
En plus des instructions générées par l'approche décrite à la section \ref{sec:instructgpt}, nous avons introduit une nouvelle tâche générative propre aux données journalistiques. L'objectif étant d'injecter le plus possible de connaissances du domaine aux modèles, les instructions sont construites de façon à produire un résumé des articles centré sur les faits principaux.
Ces résumés ont été générés automatiquement par le modèle GPT-4o-mini d'OpenAI \footnote{version gpt-4o-mini-2024-07-18} et respectent un format JSON contraint par le schéma décrit dans l'annexe \ref{annex:pydantic_of}.

\begin{promptbox}[colback=white]{Prompt pour la génération de résumés Ouest-France}
\textbf{SYSTEM}
Vous êtes un assistant AI spécialisé NLP et en extraction d'information. 
Vous devez analyser des contenus journalistiques spécialisés dans le domaine en défense et affaires militaires. 
Fournissez des informations structurées en adaptant votre langage à l'interlocuteur et en respectant le format demandé.

\textbf{USER} Lister de faits importants pour résumer l'article suivant. 
Chaque fait doit être associé à une question factuelle. Les faits et questions doivent préciser un contexte suffisant (entités, dates...) pour être compris sans lire l'article original. Formater la réponse en JSON en respectant le schéma donné.
\end{promptbox}

Cette étape a produit un total de 8\,736~instructions qui viennent s'ajouter aux instructions précédemment décrites pour la source OF, soit un total de 37\,429 instructions produites à partir des données journalistiques. Un exemple est fourni à l'annexe \ref{annex:exemples}.

\subsubsection{Amorces système}
\label{sec:amorces_systemes}
L'amorce système conditionne l'assistant dans ses réponses et permet un contrôle plus fin du comportement du modèle. 
Ainsi, les instructions générées automatiquement par le processus décrit ci-dessus se voient adjoindre en préambule une amorce système avant de les exploiter pour l'adaptation des modèles. 
Pour éviter que nos modèles adaptés ne soient trop conditionnés par des amorces systèmes répétitives et afin qu'ils gardent le même niveau de performance sur le domaine spécifique quelle que soit l'amorce utilisée par l'utilisateur, nous avons défini une procédure pour produire des amorces système variées. Ainsi, nous avons construit plusieurs amorces système de manière dynamique à l'aide d'une grammaire, comprenant plus ou moins de détails sur le comportement à adopter face à la question posée. Les descriptions utilisées dans cette grammaire ont été tout d'abord  rédigées à la main, puis réécrites et enrichies à l'aide de \texttt{Claude-3.7-Sonnet}. Au total, cette grammaire permet de générer 45 variations d'amorces systèmes qui sont ensuite associée de façon aléatoire aux instructions générées précédemment. La grammaire est fournie en détail à l'annexe \ref{annex:amorce_systeme_grammaire}.  

\subsubsection{Statistiques sur les instructions générées retenues}
L'approche décrite à la section~\ref{sec:instructgpt} a été appliquée à la fois à partir des données fournies par les organisateurs (AMIAD) et à partir des données collectées par Ouest France (OF). Au total, environ 90k instructions générées ont été sélectionnées avec cette méthode pour la source AMIAD (et seront référées par la suite par la dénomination \textbf{Gen AMIAD}) et 29k instructions pour la source OF. Une tâche spécifique a été définie de surcroît pour les données journalistiques (cf. section \ref{sec:instructof}). L'ensemble des instructions générées à partir des données journalistiques sera nommé par la suite \textbf{Gen OF}. Au final la répartition des instructions générées par type de tâche est fournie dans la table \ref{tab:instruct-tasks}. Des exemples sont fournis en annexe \ref{annex:exemples}.

\begin{table}[h!]
    \centering
    \begin{tabular}{c|cr}
         \textbf{Type d'instructions} & \textbf{Gen AMIAD} & \textbf{Gen OF}~~  \\
         \hline
         Questions factuelles & 23\,767 &  7\,447\\
         QCM & 23\,991 &  7\,416\\
         Titrage & 24\,900 & 7\,806 \\
         Résumé & 17\,935 & 6\,024 \\
         Résumé factuel & -- & 8\,736 \\ 
         Total & 90\,593 &  37\,429 \\
    \end{tabular}
    \caption{Répartition par type de tâche des instructions générées}
    \label{tab:instruct-tasks}
\end{table}

\subsection{Instructions construites à partir de patrons}
\label{sec:instructpatrons}

Pour certaines tâches comme l'explication d'acronymes ou encore les tâches de traduction, l'intérêt d'utiliser des modèles génératifs pour obtenir des instructions est moindre. Ainsi, afin de réduire notre empreinte carbone, nous utilisons des patrons d'instructions, nous permettant de construire des instructions relativement diverses dans leur formulation.
Ces patrons sont disponibles en annexes \ref{annex:patrons} et leur application a permis de produire un total de 2\,786~instructions focalisées sur la connaissance du vocabulaire métier (qui seront par la suite référencées par la dénomination \textbf{Patrons AMIAD}).

\subsection{Construction d'instructions longues}
\label{sec:instructlong}
Jusqu'ici, les instructions générées étaient construites pour répondre aux quatre tâches de NLP sus-mentionnées que l'on sait évaluer (résumé, titrage, QCM et question-réponse factuelle). En complément, afin que nos modèles continuent d'assimiler les connaissances des documents, nous avons également créé des instructions plus longues et génératives.

Pour cela, en s'appuyant sur les documents, nous avons construit de nouvelles tâches de génération, appelées instructions longues, où l'idée est de combiner, ou d'inverser certaines tâches pour passer à un problème où la réponse attendue est un texte long.

Nous avons ainsi créé les instructions génératives suivantes : 
\begin{itemize}
    \item combinaison d'acronymes : explication de 2 à 30 acronymes choisis aléatoirement  ;
    \item combinaison de traductions : traduction de 2 à 30 textes choisis aléatoirement ;
    \item résumé inversé : génération d'un paragraphe complet à partir d'un résumé.
\end{itemize}

\vspace{0.5em}
Finalement, ce sont près de 7\,800 instructions qui viennent s'ajouter au jeu d'instructions global (référencées par la suite comme \textbf{Long AMIAD}). Ces dernières diversifient nos tâches tout en permettant un nouveau passage du modèle sur les documents fournis.

\subsection{Instructions généralistes}
\label{sec:instructtulu}
Pour éviter que le modèle adapté ne perde certaines compétences préalables, comme des connaissances générales ou la capacité à respecter le format d'une instruction, nous avons également considéré l'inclusion de données hors du domaine visé. Pour cela, nous nous sommes appuyés sur le corpus \texttt{Tulu 3 SFT mixture}\footnote{\url{https://huggingface.co/datasets/allenai/tulu-3-sft-mixture}} récemment proposé dans le cadre de l'apprentissage des modèles Tülu~3~\cite{lambert2024tulu3}. Ce jeu de données est multi-tâches, multilingue et très volumineux (environ 1 million d'instructions).

Pour ne pas déséquilibrer nos données d'apprentissage avec trop d'instructions généralistes et dans la perspective de produire un modèle principalement pour le français, nous avons produit une version filtrée du corpus \texttt{Tulu 3 SFT mixture}, de sorte à n'inclure que des instructions dont la langue principale est le français\footnote{Par \og{}langue principale\fg{}, il faut comprendre que les instructions peuvent contenir des passages dans une autre langue, par exemple dans le cas d'une tâche de traduction.}. Le filtrage a été effectué \textit{via} l'outil LangDetect\footnote{\url{https://pypi.org/project/langdetect}}. La version filtrée ainsi obtenu comporte $5\,891$~instructions (référencées par la suite comme \textbf{Tulü 3 Fr}).

\subsection{Bilan}

La table \ref{tab:stats_instructions} regroupe l'ensemble des instructions produites pour l'adaptation des modèles. Celles ayant été produites à partir des segments de \texttt{train} seront utilisées pour l'affinage sur instructions décrit à la section \ref{sec:ift}, avec plusieurs configurations qui seront détaillées dans la partie expérimentale.


\begin{table}[h!]
    \centering
    \begin{tabular}{c|rr|c|rr}
                         & \multicolumn{2}{c|}{\textbf{Nb. d'instructions }} & \textbf{Nb. tokens} & \multicolumn{2}{c}{\textbf{Longueur moyenne (tokens)}}\\
         & Total~ & Apprentissage &  & \multicolumn{1}{c}{Entrée} & \multicolumn{1}{c}{Sortie}\\
         &  &                                       &  & \multicolumn{1}{c}{(utilisateur)} & \multicolumn{1}{c}{(assistant)}\\ \hline
         Gen AMIAD & 90\,593& 79\,974~~~~ & 60.9M & 577 & 93\\
         Patrons AMIAD & 2\,786 & 2\,228~~~~ & 138k & 39 & 11\\
         Long AMIAD & 7\,786 & 6\,082~~~~ & 16.8M  & 272 & 2\,008\\
         Gen OF & 37\,429 & 32\,912~~~~ & 28M & 648 & 102\\
         Tülu 3 Fr & 5\,891 & 5\,891~~~~ & 10M & 744 & 946 \\
    \end{tabular}
    \caption{Statistiques sur les différents jeux d'instructions.}
    \label{tab:stats_instructions}
\end{table}

%% file: 3_finetuning.tex
\section{Adaptation des modèles}
\label{sec:adapt}

Différentes techniques d'adaptation existent quand on souhaite spécifier et affiner les connaissances et aptitudes d'un modèle. D'abord, la poursuite du pré-entraînement (\cpt, \textit{continued pre-training}) est une méthode visant à prolonger l'apprentissage du modèle sur un nouveau dataset. Ensuite, l'Affinage Sur Instructions (ASI, \textit{instruction-tuning}) nous permet d'apprendre des nouvelles tâches à notre modèle.
Ces deux techniques, appliquées au modèle \doc{Mistral-7B-Instruct-v0.3}, sont détaillées dans les paragraphes suivants.

\subsection{Environnement logiciel et matériel}

Le \cpt{} et l'\ift{} ont été mis en œuvre avec les librairies \texttt{trl}\footnote{\url{https://huggingface.co/docs/trl/}} (version \texttt{0.15.2}), notamment la classe \texttt{trl.SFTTrainer} et \texttt{transformers}\footnote{\url{https://huggingface.co/docs/transformers}} (version \texttt{4.49.0}) de Hugging Face. Les expériences ont été menées en utilisant l'outil \texttt{deepspeed}\footnote{\url{https://www.deepspeed.ai/}} (version \texttt{0.15.3}) et en particulier, la technique ZeRO (\textit{Zero Redundancy Optimizer})~\cite{rajbhandari:2020} a été utilisée au niveau 3, qui permet de partitionner les états de l'optimiseur, les gradients et les paramètres du modèle et de décharger la mémoire des GPUs (Graphics Processing Unit) vers celle du CPU (Central Processing Unit). Le fichier de configuration de l'outil deepspeed est fourni à l'annexe~\ref{subsubsec:annexe_hyper_param} Les librairies d'apprentissage utilisent la librairie \texttt{torch}\footnote{\url{https://pytorch.org/}} (version 2.5.1) 
\texttt{cuda}\footnote{\url{https://developer.nvidia.com/cuda-toolkit}} (version 12.2.r12.2). Le calcul de l'attention est optimisée par la librairie \texttt{flash-attn}\footnote{\url{https://pypi.org/project/flash-attn/}} (version 2.7.4.post1) en utilisant l'algorithme FlashAttention-2~\cite{dao_flashattention-2:2023}. La supervision de la fonction de perte est effectuée avec l'outil \texttt{tensorboard}\footnote{\url{https://www.tensorflow.org/tensorboard}}.
Les adaptations ont été réalisées sur une carte NVIDIA H200.

Par ailleurs, à la fois pour le \cpt{} et pour l'\ift, il a fallu ajouter un token de rembourrage (\textit{padding token}) qui n'était pas présent dans le modèle \texttt{Mistral-7B-Instruct-v0.3}. Ce token qui doit être différent du token de fin de séquence est nécessaire pour pouvoir concaténer plusieurs exemples jusqu'à la taille de la fenêtre du modèle, comme cela est expliqué dans la section sur le poursuite du pré-entraînement~\ref{subsec:CPT}. Ce token de rembourrage a été ajouté grâce au code Python présenté par l'algorithme~\ref{alg:rembourrage}.

\begin{code}[t]
\begin{lstlisting}[style=python]
from transformers import AutoModelForCausalLM, AutoTokenizer 
model = AutoModelForCausalLM.from_pretrained("mistralai/Mistral-7B-Instruct-v0.3")
tokenizer = AutoTokenizer.from_pretrained("mistralai/Mistral-7B-Instruct-v0.3")
# activation du calcul du gradient pour la couche d'entree
model.enable_input_require_grads()
# ajout du token de rembourrage
tokenizer.add_special_tokens({"pad_token": "[PAD]"})
# mise a jour du nombre de plongements
model.resize_token_embeddings(len(tokenizer))
model.config.pad_token_id =\
tokenizer.convert_tokens_to_ids(tokenizer.pad_token)
\end{lstlisting}
\vspace{-2mm}
\caption{\label{alg:rembourrage}Code Python d'ajout d'un token de rembourrage}
\end{code}

Dans le code présenté, pour le chargement du modèle et du tokenizer, certains paramètres ont été omis tels que le token d'identification sur le site de téléchargement et ceux nécessaires à la librairie \texttt{flash-attn}.
Cet environnement commun est spécialisé pour chacun des deux types d'adaptation, à savoir le \cpt{} et l'\ift. 

\subsection{Poursuite du pré-entraînement (\cpt)}\label{subsec:CPT}

Le domaine visé étant technique de par son vocabulaire et ses acronymes propres, il nous semblait intéressant de tester une approche de poursuite du pré-entraînement afin que le modèle assimile une première fois les documents du challenge.

De cette façon, nous distinguons deux jeux de données différentes : 
\begin{itemize}
    \item Jeu fermé : Documents constitués à partir des données fournies par les organisateurs.
    \item Jeu ouvert : Documents constitués à partir des données fournies par les organisateurs + les documents provenant de Ouest-France.
\end{itemize}

A cette étape, l'ensemble des données fournies par l'AMIAD sont utilisées, et ce, directement après les traitements décrits dans la section \ref{sec:data}. Les hyper-paramètres de l'adaptation sont les paramètres d'initialisation de la classe \texttt{trl.SFTConfig}\footnote{\url{https://huggingface.co/docs/trl/main/en/sft_trainer\#trl.SFTConfig}}, l'algorithme~\ref{alg:hyperparameters} donne les valeurs des principaux hyper-paramètres.

\begin{code}[t]
\begin{lstlisting}[style=yaml]
gradient_accumulation_steps: 64
gradient_checkpointing: true
group_by_length: false
learning_rate: 2.0e-05
lr_scheduler_type: cosine #learning rate scheduler type
max_grad_norm: 1.0 # maximum gradient norm
num_train_epochs: 3 # number of training epochs 
optim: paged_adamw_32bit
per_device_train_batch_size: 2 # per device training batch size
warmup_ratio: 0.05
weight_decay: 0.1
\end{lstlisting}
\vspace{-5mm}
\caption{\label{alg:hyperparameters}Hyper-paramètres}
\end{code}

La définition de l'ensemble des paramètres est donnée dans un fichier au format \texttt{yaml} présenté à l'annexe~\ref{subsubsec:annexe_hyper_param}. Les paramètres qui ne sont pas définis dans l'annexe prennent les valeurs par défaut définis par la fonction d'initialisation de \texttt{trl.SFTConfig}, à l'exception du paramètre \texttt{packing} qui est initialisé à \texttt{True} et du paramètre \texttt{max\_seq\_length} qui est initialisé à la taille de la fenêtre du modèle. Le paramètre \texttt{packing} permet de concaténer plusieurs exemples à la suite dans un seul exemple jusqu'au maximum de \texttt{max\_seq\_length} caractères. Cette concaténation est réalisée par la fonction \texttt{\_prepare\_dataset} de la classe \texttt{trl.SFTTrainer}\footnote{\url{https://huggingface.co/docs/trl/main/en/sft_trainer\#trl.SFTTrainer}}. Les exemples concaténés sont alors les considérés pour la constitutions des lots lors de l'apprentissage.

Partant d'un modèle instruit (à savoir \texttt{Mistral-7B-Instruct-v0.3}), la pertinence \textit{a priori} de reprendre une étape de pré-entraînement pouvait sembler questionnable mais la courbe de la fonction de perte de la figure~\ref{fig:loss_curve_cpt} permet d'observer que la valeur de perte finale est inférieure à celle initiale. L'évolution est cependant marquée par une augmentation lors d'une première phase. Elle peut sans doute s'expliquer par le fait que le modèle s'attend à recevoir des instructions, plutôt que des documents. Dans une seconde phase, la courbe diminue progressivement pour finalement se stabiliser. Après une période de confusion, le modèle interprète correctement les documents qui lui sont fournis. Il gagne alors peu à peu en précision ce qui se matérialise par une convergence progressive de la fonction de perte. En outre, si la valeur finale de la perte n'est pas si éloignée de son point initial~--~bien que quand même inférieure, il est intéressant de noter que de nouvelles données spécifiques au domaine ont alors été vues par le modèle, ce qui est prometteur pour la suite des expérimentations.

Sur un autre plan, les premiers apprentissages ont montré que les modèles résultant de la poursuite du pré-entraînement ne produisaient jamais le symbole de fin de séquence \texttt{EOS} (\textit{End Of Sequence}) et donc généraient de nouveaux tokens indéfiniment. Ce symbole aurait dû être rajouté à chaque document par la méthode \texttt{\_prepare\_dataset} de la classe \texttt{trl.SFTTrainer} mais ce n'était pas le cas pour la version \texttt{0.15.2} de la bibliothèque de \texttt{trl} utilisée lors des expériences\footnote{Ce défaut a été corrigé dans les versions plus récentes}. Nous avons alors explicitement ajouté le symbole de fin de séquence à la fin de chaque document lors du chargement des données d'adaptation et de validation. Cette modification a permis de rétablir la prédiction  de la fin de séquence et d'arrêter la génération du texte lors de l'utilisation des modèles.

\begin{figure}[t!]
\centering
\includegraphics[width=\textwidth]{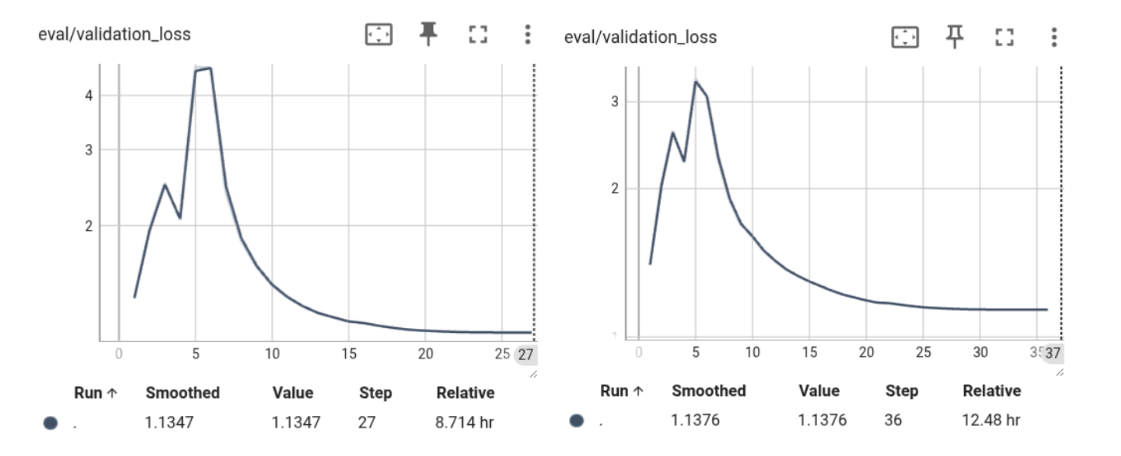}
\caption{Courbe de la fonction de perte du \cpt{} pour les jeux fermé et ouvert. Les courbes affichées sont régularisées et proviennent des fichiers de sortie de \textit{TensorBoard}.}
\label{fig:loss_curve_cpt}
\end{figure}

\subsection{Affinage sur instructions (\ift)}
\label{sec:ift}
Tout comme pour le pre-entraînement, nous avons continué l'entraînement de nos modèles sur nos corpus d'instructions via le \texttt{trl.SFTrainer} de la librairie \texttt{transformers}.
Chaque instruction est représentée par une liste de 3 dictionnaires. Chaque dictionnaire a deux clés:
\begin{itemize}
    \item \texttt{"role"}: la valeur précise le rôle attribué au texte contenu dans le champ \texttt{"content"}, elle peut prendre l'une des valeurs suivantes \texttt{"system"}, \texttt{"user"}, ou \texttt{"agent"},
    \item \texttt{"content"}: la valeur contient le texte associé à chacun des rôles.
\end{itemize}

Le premier dictionnaire contient le \texttt{"role"} \texttt{"system"}, le deuxième le \texttt{"role"} \texttt{"user"} et le dernier le \texttt{"role"} \texttt{"agent"}. Ils correspondent respectivement au rôle que doit jouer le système, à la question ou la requête formulée par l'utilisateur et à la réponse fournie par l'agent. Des exemples de ce format sont montrés sur le site la documentation d'application de modèle de conversation \footnote{\url{https://huggingface.co/docs/transformers/en/chat\_templating}}. Cette liste est transformée en une chaîne de caractères par la fonction \texttt{transformers.PreTrainedTokenizerBase.apply\_chat\_template}\footnote{\url{https://huggingface.co/docs/transformers/internal/tokenization\_utils\#transformers.PreTrainedTokenizerBase.apply\_chat\_template}} qui introduit les séparateurs textuels spécifiques à chaque tokenizer entre les différents éléments de l'instruction (c'est-à-dire les 3 élements de la liste initiale).
\\
De plus lors de l'\ift, l'optimisation de la fonction de perte ne se fait que sur la partie du texte correspondant au rôle \texttt{"agent"}. La fonction de préparation des lots d'exemples \texttt{trl.DataCollatorForCompletionOnlyLM} permet ce calcul de la fonction de perte. En plus de la conversion de la chaîne de caractères en suite de tokens, elle indique sur quels tokens il ne faut pas optimiser la fonction de perte. Il faut lui fournir dans le paramètre \texttt{response\_template} les tokens correspondant aux séparateurs entre la fin de la requête de l'utilisateur et le début de la réponse de l'agent. Ces tokens sont identifiés au préalable sur une instruction factice. Le paramètre \texttt{padding\_free} positionné à \texttt{False} permet de concaténer plusieurs exemples d'un même lot\footnote{\url{https://huggingface.co/blog/packing-with-FA2}}, dans une seule fenêtre du modèle de langage, de ne calculer la fonction de perte que sur les réponses de l'agent, de respecter les bornes entre chaque exemple à l'intérieur de la fenêtre et d'utiliser les mécanisme Flash-Attention-2~\cite{dao_flashattention-2:2023}. Contrairement à ce qui est réalisé pour le pré-entraînement continué~\ref{subsec:CPT}, la concaténation est réalisée pour chaque lot et le paramètre \texttt{packing} de l'objet \texttt{trl.SFTConfig} est ici forcé à \texttt{False}.

Enfin, quelques paramètres de l'objet \texttt{trl.SFTConfig} utilisé pour réaliser l'affinage sont donnés dans l'algorithme~\ref{alg:hyperparameters_asi}.

\begin{code}[t]
\begin{lstlisting}[style=yaml]
learning_rate: 2.0e-06
lr_scheduler_type: cosine #learning rate scheduler type
num_train_epochs: 2 # number of training epochs 
optim: paged_adamw_32bit
per_device_train_batch_size: 16 # per device training batch size
warmup_ratio: 0.05
weight_decay: 0.01
\end{lstlisting}
\vspace{-5mm}
\caption{\label{alg:hyperparameters_asi}Hyper-paramètres pour l'affinage sur instructions}
\end{code}


Outre les différences sur la concaténation des exemples, par rapport à l'étape de poursuite du pré-entraînement décrite à la section~\ref{subsec:CPT}, deux hyper-paramètres ont été modifiés : le \textit{learning rate} est réduit d'un ordre de grandeur et nous entraînons le modèle pour seulement deux époques au lieu de trois. Ces changements sont motivés par des expériences d'optimisation de ces hyper-paramètres. L'ensemble des paramètres est fourni dans un fichier \texttt{yaml} présenté à l'annexe~\ref{subsubsec:annexe_hyper_param}. L'impact du \textit{learning rate} est illustré en annexe dans la figure \ref{fig:lr_comparison_it}. Les autres valeurs sont les valeurs définies par défaut pour l'initalisation de l'objet \texttt{trl.SFTConfig}.

%% file: 4_eval.tex
\section{Evaluation}
\label{sec:eval}
\subsection{Modèles retenus pour le challenge}

Nous avons soumis 3 modèles 7B dans chacune des conditions "\texttt{Fermé}" (en exploitant uniquement les données fournies par les organisateurs du challenge) et "\texttt{Ouvert}" (en exploitant également des données additionnelles). 
Les caractéristiques des 6 modèles sélectionnés sont fournies dans la table \ref{tab:runs}. Les choix ont principalement été guidés par la possibilité d'analyser l'impact de la phase de Poursuite du du Pré-Entrainement (\cpt) d'une part et l'impact de la diversité des instructions pour l'affinage (\ift) d'autre part. 

En ce qui concerne les modèles en condition \texttt{Fermé}, nous avons choisi de soumettre deux modèles ayant uniquement subi un affinage sur instructions. Le \texttt{run1} a été adapté uniquement sur les instructions de type \textit{Résumé}, \textit{Titrage}, \textit{Question factuelle}, et \textit{QCM} générées à partir des données AMIAD selon le processus décrit à la section \ref{sec:instructgpt}. Le \texttt{run2} a été adapté avec les instructions complémentaires produites à l'aides de Patrons sur deux documents spécifiques (cf section \ref{sec:instructpatrons} ainsi qu'à l'aide d'instructions longues (cf section \ref{sec:instructlong}). L'objectif ici était de mesure si des connaissances du domaine pouvaient être apportées au modèle uniquement sur la base d'instructions. Le \texttt{run3} correspond à une étape préliminaire de \cpt. 

Concernant les modèles en condition \texttt{Ouvert}, nous avons systématiquement appliqué l'étape de \cpt  sur l'ensemble des données fournies par les organisateurs (AMIAD) et par Ouest France (OF). Le \texttt{run1} a été affiné à l'aide d'instructions produites uniquement à partir des données AMIAD alors que le \texttt{run2} a également été affiné à l'aide d'instructions produites sur la base des données OF. Enfin, pour le \texttt{run3}, un jeu d'instructions généralistes en français (cf section \ref{sec:instructtulu}).

\begin{table}[]
\begin{tabular}{l|c@{~~}c|ccccc}
 & \multicolumn{2}{c|}{\textbf{Documents \cpt }} & \multicolumn{5}{c}{\textbf{Instructions \ift }} \\
 &  &  &      Gen     &  Patrons & Long &  Gen  &  Tülu 3\\
 & AMIAD & OF &      AMIAD     &  AMIAD & AMIAD &  OF &  Fr \\
\closedruna &           &          &  \checkmark  &   &   &   &   \\
\closedrunb &           &          &  \checkmark & \checkmark  & \checkmark  &   &    \\
\closedrunc &     \checkmark &          &  \checkmark & \checkmark  & \checkmark  &   &    \\
\openruna &     \checkmark &     \checkmark     &  \checkmark & \checkmark  & \checkmark  &   &    \\
\openrunb &      \checkmark &     \checkmark     &  \checkmark & \checkmark  & \checkmark    &   \checkmark &    \\
\openrunc &       \checkmark &     \checkmark     &  \checkmark & \checkmark  & \checkmark   &   \checkmark &   \checkmark \\
\hline
\textbf{Instr. d'apprentissage} & -- & -- & 79\,974 & 2\,228 & 6\,082 & 32\,912 & 5\,891 \\
\textbf{Nb. total de tokens} & 48.6M & 11.5M & 53.5M & 110k & 13.6M & 24.3M & 10M
\end{tabular}
\caption{Description des configurations pour les différents modèles soumis. Détail des données sélectionnées pour les phases de \cpt~et d'\ift, à partir des données fournies par les organisateurs du challenge (AMIAD) et par Ouest-France (OF).}
\label{tab:runs}
\end{table}

\subsection{Protocole d'évaluation}
Afin d'évaluer la qualité des différents modèles entraînés, nous nous appuyons sur plusieurs jeux de données complémentaires et diversifiés. 
Dans la suite de cette section nous décrivons les jeux de données de test utilisés pour l'évaluation, ainsi que les métriques utilisées. Lorsque les tâches sont évaluées par une métrique d'\textit{accuracy}, nous décrivons le détail de l'évaluation de l'exactitude de la réponse. Pour les tâches génératives, plusieurs métriques ont été envisagées mais nous avons choisi de reporter une métrique de type MOS. Un score moyen d’opinion est estimé sur l’ensemble des échantillons, par une note de 0 à 5 qui est attribuée pour chaque échantillon par le modèle \texttt{GPT-4o-mini} selon le principe du \textit{LLM As a Judge}, indiquant la pertinence du résumé généré par rapport à la référence. Ces aspects sont bien évidemment discutables mais dans le cadre de ce challenge, nous avons dû opter pour une approche d'évaluation pragmatique pour guider les développements et sélectionner les modèles à retenir dans un temps contraint.

\subsection{Évaluation sur des jeux de données du domaine}

Nous avons évalué nos modèles sur trois types de jeux de test~: le premier est dérivé des ensembles d'instructions synthétiques générées à partir des documents (cf.section~\ref{eval:synt}), le deuxième provient des organisateurs du challenge (cf. section~\ref{eval:amiad}) et le dernier regroupe plusieurs évaluations sur des données génériques (cf. section~\ref{eval:gene}).

\subsubsection{Benchmarks synthétiques} 
\label{eval:synt}
Nous avons retenu pour l'évaluation une partie des instructions décrites à la section \ref{sec:instruct} qui avaient été écartées dans la partition de test. De façon à ne pas trop surcharger nos jeux de tests, nous avons réalisé un échantillonnage aléatoire dans chacun des ensembles d'instructions pour constituer une base de test issues des différentes sources de données. Nous n'avons pas fait d'évaluation sur les instructions générées à partir des documents journalistiques, mais uniquement sur celle générées à partir des documents fournis par les organisateurs.
\begin{itemize}
    \item \textbf{QCM} regroupe des échantillons de questions à choix multiples générés à l'aide de GPT-4.1-mini tels que décrits à la section \ref{sec:instructgpt}. Ils sont évalués à l'aide d'une métrique d'\textit{accuracy} sur la lettre extraite de la réponse (l'extraction de la lettre est réalisée à l'aide de GPT-4o-mini).
    \item \textbf{Questions factuelles}  regroupe des échantillons de questions à choix multiples générés à l'aide de GPT-4.1-mini tels que décrits à la section \ref{sec:instructgpt}. L'évaluation est faite à l'aide d'une métrique d'accuracy, où une réponse générée est considérée comme correcte si elle contient la chaîne de caractère correspondant au fait de référence. Plusieurs variantes d'écriture pour le fait de référence sont acceptées : pour les dates (JJ/MM/AAAA, MM/AAAA, JJ/MM, formats textuels, abréviations etc.), les nombres (séparateurs, suffixes d'unités, gestion des signes, formats mixtes etc..). Le texte est également normalisé afin de traiter correctement la ponctuation et les caractères de contrôle.
    \item \textbf{Acronymes} correspond à un échantillonage de la partition de test pour le jeu de données construit à l'aide de patrons (cf. section \ref{sec:instructpatrons}. La métrique utilisée est l'\textit{accuracy}, où une réponse générée est considérée comme correcte si elle contient la signification de référence de l'acronyme. La réponse générée ainsi que la référence sont tous les deux normalisés au préalable (passage des textes en minuscules, conversion des caractères unicode en ASCII, gestion des espaces) afin de rendre l'évaluation plus souple. Dans le cas où plusieurs significations existent pour un même acronyme, la réponse générée est considérée correcte si au moins une de ces significations est présente dans la réponse.
    \item \textbf{Résumé} contient un échantillonnage aléatoire de la partition de test des instructions générées selon l'approche décrite à la section \ref{sec:instructgpt}. Pour limiter le coût de l'évaluation, nous n'avons conservé que les instructions produites à partir des sources\textit{ web-minarm-adt}, \textit{web-minarm-airinterne-divers}, \textit{interne-ema} et \textit{wikipedia-portail-armee}. La métrique utilisée est la métrique MOS décrite en début de cette section.
    \item \textbf{Titrage} contient un échantillonnage aléatoire de la partition de test des instructions générées selon l'approche décrite à la section \ref{sec:instructgpt}. Pour limiter le coût de l'évaluation, nous n'avons conservé que les instructions produites à partir des sources\textit{web-minarm-dga}, \textit{web-minarm-marine}, \textit{interne-formation}, \textit{interne-magazines} et \textit{wikipedia-portail-armee}. La métrique utilisée est la métrique MOS décrite en début de cette section. 
\end{itemize}

Les résultats obtenus sur ces jeux de données sont fournis à la table \ref{tab:evalgen}.

\begin{table}[h!]
\centering
\begin{tabular}{@{}l@{~~}c@{~~}c@{~~}c@{~~}c@{~~}c@{~~}|@{~~}c}
& \textbf{Résumé} & \textbf{Titrage} & \textbf{QCM} & \textbf{Q. fact.} & \textbf{Acron.} & \textbf{Rg. moy.} \\
\hline
\textbf{Nombre d'échantillons} & 620 & 542 & 2\,083 & 2\,029 & 128 & \\
\textbf{Métrique} & MOS & MOS & Accuracy & Accuracy & Accuracy & \\
\hline

\texttt{GPT-4o-mini}  (2024-07-18) & 3,83\rang{3} & 3,38\rang{3} & 69,1\rang{1} & 9,6\rang{5} & 10,9\rang{4} & 3,2 \\
\texttt{Mist.-Sm.-24B-Inst.-2501} & 3,55\rang{5} & 3,26\rang{5} & 68,1\rang{2} & 6,5\rang{8} & 7,0\rang{7} & 5,4 \\
\hline
\texttt{Mistral-7B-Instruct-v0.3} & \textit{3,24}\rang{6} & \textit{3,27}\rang{4} & \textit{57,5}\rang{9} & \textit{5,6}\rang{9} & \textit{3,9}\rang{8} & 7,2 \\
\hline
\closedruna & \textbf{3,93}\rang{1} & \textbf{3,81}\rang{1} & 65,4\rang{5} & 9,2\rang{6} & 3,9\rang{8} & 4,2 \\
\closedrunb & 3,86\rang{2} & 3,76\rang{2} & 64,4\rang{7} & 6,8\rang{7} & 8,6\rang{6} & 4,8 \\
\closedrunc & 2,10\rang{9} & 2,06\rang{9} & 62,2\rang{8} & 10,3\rang{4} & 10,9\rang{5} & 7,0 \\
\hline
\openruna & 2,66\rang{7} & 2,32\rang{7} & 65,5\rang{4} & \textbf{13,2}\rang{1} & \textbf{15,6}\rang{1} & 4,0 \\
\openrunb & 2,58\rang{8} & 2,22\rang{8} & 65,1\rang{6} & 12,0\rang{2} & 14,8\rang{2} & 5,2 \\
\openrunc & 3,66\rang{4} & 3,25\rang{6} & \textbf{65,6}\rang{3} & 11,3\rang{3} & 14,8\rang{2} & \textbf{3,6} \\

\hline
\end{tabular}
\caption{Résumé des résultats sur les tâches générées à partir des données AMIAD. En gras, les meilleurs résultats parmi les modèles \texttt{Mistral-7B-Instruct-v0.3} et ses versions adaptées.}
\label{tab:evalgen}
\end{table}

De façon générale, les modèles ayant subi une poursuite du pré-entrainement (\cpt) sont moins performants sur les tâches génératives de résumé et de titrage, tout du moins du point de vue de la métrique MOS implémentée en mode \textit{LLM-as-a-judge}. En effet dans certains cas, les modèles 
avec \cpt{} ont tendance à produire des sorties plus longues ou à boucler sur leurs réponses. Seule la version \openrunc{} en condition ouverte permet de retrouver des performances acceptables, sans doute grâce à l'apport des instructions généralistes issues du corpus Tülu. Les modèles \closedruna{} et \closedrunb, qui n'ont pas subi de \cpt, sont quant à eux performants sur les tâches de résumé et de titrage, outrepassant les performances des baselines. Ce résultat est à relativiser par le fait que les instructions de test ont été générées selon des protocoles identiques aux instructions d'apprentissage. Dans ce contexte, la phase d'\ift{} s'apparente à un affinage à la tâche.

Concernant la tâche de QCM, il est intéressant de voir que les modèles baselines les plus performants évalués ici, n'ont pas des performances très importantes (inférieures à 70\%), ce qui reflète bien la problématique des connaissances spécifiques. Le modèle 7B de référence \texttt{Mistral-7B-Instruct-v0.3} présente quant à lui des performances inférieures avec une accuracy de 57,5\%. Pour cette tâche les meilleurs modèles adaptés à partir de ce modèle 7B atteignent des performances de l'ordre de 65\%. Dans le cas des données fermées, le run \closedrunc~qui a subi une poursuite de pré-apprentissage présente des performances inférieures. Une analyse plus détaillée des sorties montre que cette dégradation peut être également expliquée par des dysfonctionnements de format, avec des sorties longues et en boucle, rendant plus difficiles l'exploitation des sorties. 

Enfin les tâches de question factuelles et d'acronymes présentent globalement des performances très basses, y compris pour les modèles \textit{baseline}, même si l'adaptation produit une amélioration notable, en particulier pour le modèle \closedruna. Il est intéressant de noter que la seule différence entre \openrunc~et \closedruna~est l'ajout des données journalistiques Ouest France lors de la phase de poursuite du pré-apprentissage. L'écart de performances observées, sur des tâches qui ne sont pas issues du contexte journalistique laisse entrevoir que de telles données apportent une meilleure connaissance factuelle du domaine. Malgré cela, il reste des pistes à explorer pour améliorer l'injection de connaissances factuelles.

\subsubsection{Benchmarks fournis par les organisateurs}
\label{eval:amiad}
Quelques échantillons de test ont été fournis, que nous avons également intégré pour l'évaluation, même si les résultats doivent être pris avec les précautions d'usage du fait de leur petite taille.
\begin{itemize}
    \item \textbf{QCM Défense} regroupe les différents jeux de données de type QCM fournis sur les domaines de la défense (QCM Culture SecuDefense, QCM Culture Minarm, QCM Institution FR). Ils sont évalués à l'aide d'une métrique d'\textit{accuracy} sur la lettre extraite de la réponse (l'extraction de la lettre est réalisée à l'aide de GPT-4o-mini).
    \item \textbf{Gold} contient des exemples de questions factuelles, dont la réponse de référence est fournie sous la forme d'une expression régulière. Ils sont évalués à l'aide d'une métrique d'\textit{accuracy} sur le respect de l'expression régulière de référence dans la réponse générée par les modèles.
    \item \textbf{Résumé court}\footnote{Pour des raisons matérielles et de temps nous n'avons pas inclus les résumés longs fournis par les organisateurs dans nos évaluations.} contient des exemples de résumés de documents restreints à deux phrases en sortie. Ils sont évalués à l'aide de la métrique MOS décrite précédemment.
\end{itemize}

Le jeu de données "hallucination" a été écarté de l'évaluation, en raison des limites de sa méthode de validation automatique, qui repose uniquement sur la détection de motifs par regexp et où des réponses correctes pouvaient être considérées à tort comme des hallucinations. 

Les résultats obtenus sur ces jeux de données sont fournis à la table \ref{tab:evalamiad}.

\begin{table}[h!]
\centering
\begin{tabular}{lccc|c}
& \textbf{QCM Défense} & \textbf{Gold} & \textbf{Résumé Court} & \textbf{Rg. moy.} \\
\hline
\textbf{Nombre d'échantillons} & 30 & 10 & 10 & \\
\textbf{Métrique} & Accuracy & Accuracy & MOS \\
\hline
\texttt{GPT-4o-mini}  & 66,7\rang{1} & 40,0\rang{5} & 3,40\rang{1} & 2,3 \\
\texttt{Mist.-Sm.-24B-Inst.-2501} & 60,0\rang{3} & 20,0\rang{7} & 3,20\rang{2} & 4,0 \\
\hline
\texttt{Mistral-7B-Instruct-v0.3} & \textit{43,3}\rang{9} & \textit{10,0}\rang{9} & \textit{2,90}\rang{6} & 8,0 \\
\hline
\closedruna & 50,0\rang{7} & 20,0\rang{7} & \textbf{3,20}\rang{2} & 5,3 \\
\closedrunb & 50,0\rang{7} & 30,0\rang{6} & \textbf{3,20}\rang{2} & 5,0 \\
\closedrunc & 56,7\rang{4} & 50,0\rang{3} & 2,10\rang{9} & 5,3 \\
\hline
\openruna & 56,7\rang{4} & \textbf{60,0}\rang{1} & 2,40\rang{8} & 4,3 \\
\openrunb & \textbf{63,3}\rang{2} & \textbf{60,0}\rang{1} & 2,90\rang{6} & \textbf{3,0} \\
\openrunc & 53,3\rang{6} & 50,0\rang{3} & \textbf{3,20}\rang{2} & 3,7 \\

\hline
\end{tabular}
\caption{Résumé des résultats sur les tâches provenant des échantillons fournis par les organisateurs. En gras, les meilleurs résultats parmi les modèles \texttt{Mistral-7B-Instruct-v0.3} et ses versions adaptées.}
\label{tab:evalamiad}
\end{table}

Même si ces résultats sont à prendre avec beaucoup de précautions du fait du faible nombre d'échantillons, ils révèlent une tendance qui confirment les résultats précédents, à savoir que la poursuite du pré-entraînement permet d'améliorer l'apport de connaissances pour le domaine et que l'ajout d'instructions généralistes dans la phase d'affinage sur instructions permet de meilleurs comportements sur les tâches génératives.

\subsubsection{Jeux de données généralistes}
\label{eval:gene}
Afin d’évaluer les capacités généralistes des modèles, et notamment de vérifier l'absence de régression des performances après l'affinage, nous avons introduit des jeux de données de test généralistes dans nos évaluations. 
\begin{itemize}
\item \textbf{QCM Généraliste} regroupe les différents jeux de données de type QCM fournis par les organisateurs sur les connaissances générales (QCM culture générale, QCM égalité H/F, QCM numérique, QCM raisonnement).
\item \textbf{MMLU} correspond à une sélection de questions à choix multiples issus du corpus MMLU~\cite{hendrycks2020measuring} dans sa version anglaise (/cais/mmlu\footnote{\url{https://huggingface.co/datasets/cais/mmlu}}) et française (/openai/MMMLU\footnote{\url{https://huggingface.co/datasets/openai/MMMLU}}),  Nous avons sélectionné 10 échantillons par thématique, soit un total de 570 questions à choix multiples par langue.
\item \textbf{IFEval} : Utilisation des versions anglaise et française depuis le corpus M-IFEval~\cite{dussolle-etal-2025-ifeval}, afin d’évaluer les capacités des modèles à respecter des contraintes de format. La version anglaise contient 541 échantillons alors que la française en contient 235. La métrique retenue dans nos tableaux traduit la capacité des modèles à satisfaire toutes les contraintes donnée pour un prompt donnée, sans tolérance~\cite{dussolle-etal-2025-ifeval}. Celle-ci est notée \og{}Accuracy (stricte)\fg{}.
\end{itemize}

Les résultats sur ces jeux de données de test généralistes sont fournis dans la table~\ref{tab:resultats_benchmarks_generalistes}.
 
\begin{table}[h!]
\centering
\begin{tabular}{@{}l@{~~}c@{~~}c@{~~}c@{~~}c@{~~}c|c}
& \textbf{QCM} & \textbf{MMLU} & \textbf{MMLU} & \textbf{IFEval} & \textbf{IFEval} & \textbf{Rg.} \\
& \textbf{Généraliste} & \textbf{En} & \textbf{Fr} & \textbf{En} & \textbf{Fr} & \textbf{moy.} \\
\hline
\textbf{Nombre échantillons} & 40 & 570 & 570 & 541 & 235 & \\
\textbf{Métrique} & Accuracy & Accuracy & Accuracy & Accuracy & Accuracy & \\
& & & & (stricte) & (stricte) \\
\hline
\texttt{GPT-4o-mini} & 80,0\rang{2} & 79,8\rang{1} & 76,8\rang{1} & 80,0\rang{1} & 79,6\rang{1} & 1,2 \\
\texttt{Mist.-Sm.-24B-Inst.-2501} & 92,5\rang{1} & 72,6\rang{2} & 66,8\rang{2} & 76,9\rang{2} & 77,0\rang{2} & 1,8\\
\hline
\texttt{Mistral-7B-Instruct-v0.3} & \textit{62,5}\rang{4} & \textit{53,2}\rang{4} & \textit{48,6}\rang{4} & \textit{48,4}\rang{8} & \textit{51,9}\rang{5} & 5,0 \\
\hline
\closedruna & 60,0\rang{5} & 51,2\rang{6} & \textbf{49,5}\rang{3} & \textbf{55,5}\rang{3} & \textbf{54,0}\rang{3} & \textbf{4,0} \\
\closedrunb & \textbf{67,5}\rang{3} & \textbf{53,3}\rang{3} & 47,5\rang{5} & 54,5\rang{4} & 48,5\rang{6} & 4,2 \\
\closedrunc & 57,5\rang{7} & 48,1\rang{9} & 43,0\rang{6} & 44,5\rang{9} & 43,8\rang{7} & 7,0 \\
\hline
\openruna & 60,0\rang{5} & 51,4\rang{5} & 42,3\rang{8} & 49,7\rang{6} & 43,4\rang{8} & 6,4 \\
\openrunb & 57,5\rang{7} & 48,9\rang{7} & 42,3\rang{8} & 49,0\rang{7} & 41,3\rang{9} & 7,6\\
\openrunc & 57,5\rang{7} & 48,4\rang{8} & 42,6\rang{7} & 50,3\rang{5} & 53,6\rang{4} & 6,2\\

\hline
\end{tabular}
\caption{Résultats comparatifs des différents modèles sur les benchmarks QCM Généraliste (fourni par AMIAD), MMLU (anglais et français) et IFEval (anglais et français). En gras, les meilleurs résultats parmi les modèles \texttt{Mistral-7B-Instruct-v0.3} et ses versions adaptées.}
\label{tab:resultats_benchmarks_generalistes}
\end{table}

Les performances sur le benchmark MMLU de QCMs généralistes baissent dès lors que l'on introduit de la poursuite de pré-entraînement. En revanche, l'affinage sur instructions seul (\closedruna~et \closedrunb) n'entraîne pas de dégradation sur cette tâche, voire même entraîne une légère amélioration sur l'échantillon en français par rapport au modèle 7B de base. Sur l'échantillon de QCMs généralistes fournis par les organisateurs, le deuxième modèle \closedrunb~(affiné à l'aide des instructions générées ainsi qu'avec des instructions longues et des instructions à base de patrons issues des documents spéciaux) semble mieux se comporter, même si la petite taille du corpus incite à nuancer les observations. 

Concernant le jeu de test IFEval, qui mesure le respect des instructions (\textit{Instruction Following}), l'affinage sur le jeu d'instructions généré à partir des documents du challenge (\closedruna) apporte un gain par rapport au modèle 7B instruit initial. L'étape de poursuite du pré-entraînement induit une dégradation des performances sauf pour le run \openrunc, où les instructions Tülu 3 Fr ont été utilisées, permettant de réintroduire de la diversité dans les instructions, y compris avec des instructions de code. Ce dernier résultat confirme plusieurs analyses précédentes selon lesquelles la poursuite du pré-entraînement sur un modèle instruit peut être bénéfique pour l'adaptation au domaine, moyennant une phase ultérieure d'affinage sur instructions suffisamment riche pour compenser la perte des capacités du modèle à bien suivre les instructions.

\subsection{Evaluation du coût carbone}
La table \ref{tab:stats_carbon_consumption} fournit l'évaluation du coût carbone des différentes étapes du processus d'adaptation. Nous détaillons le coût lié aux étapes d'apprentissage (affinage par instruction (\ift) et poursuite du pré-entraînement (\cpt) le cas échéant. Par ailleurs, nous donnons également le coût lié à la génération des instructions pour l'étape d'\ift. 
La génération d'instructions a été effectuée à l'aide de \texttt{GPT-4.1-mini} sur un serveur se trouvant en Norvège. Aucun détail sur l'architecture du modèle n'ayant été partagé par OpenAI, nous utilisons la calculatrice \href{https://ecologits.ai/latest/}{EcoLogits} afin d'obtenir une estimation à partir de tokens générés pour la création du jeu d'instructions. Au total, 16 901 684 tokens \texttt{GPT-4.1-mini} ont été générés (comprend la génération de tous les champs de la sortie structurée, et non pas uniquement de l'instruction finale, pour toutes les partitions), ce qui revient à 1 890 gCO2e (65.7 kWh).

Les apprentissages pour le challenge ont été réalisés sur un cluster de 8 GPUs de modèle H200 se trouvant en France, présentant chacun 141 Go de mémoire GPU. Les valeurs d'équivalent carbone consommés par nos modèles ont été estimés via la calculatrice \textit{Green Algorithms calculator} \footnote{\href{https://calculator.green-algorithms.org/}{https://calculator.green-algorithms.org/}}. 


Enfin, comme mentionné plus tôt dans ce rapport, l'apprentissage du modèle de base a été poursuivi sur certains modèles (\cpt). De cette façon, les modèles ouverts 1, 2 et 3 ont été affinés sur instructions à partir du même modèle issu du \cpt.

\begin{table}[h!]
    \centering
    \begin{tabular}{cccc}
         \textbf{Modèle} & \textbf{\cpt} & \textbf{\ift} & \textbf{Gen. instructions} \\ \hline
         \closedruna & 0 & 385 (7.5) & 1 420 (45.9) \\
         \closedrunb & 0 & 387 (7.5) & 1 420 (45.9) \\
         \closedrunc & 724 (14.1) & 388 (7.6) & 1 420 (45.9) \\ \hline
         \openruna & 930 (18.1) & 386 (7.5) & 1 420 (45.9) \\
         \openrunb & 930 (18.1) & 560 (10.9) & 1 730 (58.9) \\
         \openrunc & 930 (18.1) & 623 (12.2) & 1 730 (58.9) \\
    \end{tabular}
     \caption{Coût carbone des modèles, en gCO2e (et en kWh). Le coût carbone associé à la génération des instructions est une estimation calculée sur la base des instructions générées par GPT-4.1-mini pour la partition de train. Le \cpt~en mode \texttt{Ouvert} n'a été réalisé qu'une fois et est commun aux trois runs.}
    \label{tab:stats_carbon_consumption}
\end{table}

L'équivalent carbone est une mesure intéressante à observer, cela dit, elle dépend très fortement de la localisation du cluster de calcul. Par exemple, toujours selon cette calculatrice, pour une même consommation énergétique, l'équivalent CO2 d'un apprentissage en Allemagne est 6.5 fois plus élevé qu'en France. C'est pour cette raison que nous indiquons également l'équivalent en kWh.


%% file: 5_annexes.tex
\appendix
\newpage

\section{Génération d'instructions}
\label{annex:instruct}

\subsection{Grammaires pour guider la génération des instructions}
\label{annex:grammar_additional_information}
Les grammaires suivantes produisent des consignes supplémentaires, spécifiques à chaque tâche, qui sont incluses dans l'amorce système lors de la génération des instructions en lieu et place du champ \texttt{additional\_information}.

\begin{promptwheelbox}[colback=white]{Consignes de génération : QCM}
\texttt{\textbf{Prompt}} -> \texttt{\textbf{InitialInstruction}} "Important instruction to follow: You will generate " \texttt{\textbf{NumberOfChoices}}\newline

\texttt{\textbf{InitialInstruction}} -> "Your question must be context-free: this means it should be answerable even without seeing the input text. Your question should concern an atomic fact and the answers should be short \& easily verifiable. The different options provided to the user must all be sensible (but all but one should be actually true). In your justification, you will not refer to the original text."\newline
\texttt{\textbf{NumberOfChoices}} -> "4 different choices: a, b, c, d." | "5 different choices: a, b, c, d, e."
\end{promptwheelbox}

\begin{promptwheelbox}[colback=white]{Consignes de génération : Résumé}
\texttt{\textbf{Prompt}} -> \texttt{\textbf{Context}} | ""\newline
\texttt{\textbf{Context}} -> "You will ask the user a summarization question. Your summarization question must only ask the user to summarize and not mention the subject of the text."
\end{promptwheelbox}

\begin{promptwheelbox}[colback=white]{Consignes de génération : Titrage}
\texttt{\textbf{Prompt}} -> "The question asked must follow this constraint: " \texttt{\textbf{constraint}}\newline
\texttt{\textbf{constraint}} -> "It must be short and to the point ('Titrer ce document')" | "It must be short and to the point ('Générer un titre pertinent pour le document-ci dessus' or slight alternatives)" | "You must specify a certain titling style" | "You must specify that the title should be suitable for a document of type [YOUR CHOICE]" | "You must specify a target style for the title and that the title should be suitable for a document of type [YOUR CHOICE]"
\end{promptwheelbox}

Il n'y a pas de consignes supplémentaires à suivre pour générer des instructions de réponse à des questions factuelles.

\subsection{Description des classes Pydantic pour le contrôle du format}
\label{annex:pydantic_classes}
Les classes Pydantic permettent de forcer la génération à suivre un certain format. La réponse générée par \texttt{GPT-4.1-mini} est donc structurée en différents champs, qui sont écrit en gras ci-dessous.

\subsubsection{Résumé}
\input{pydantic_summarization}

\subsubsection{Titrage}
\input{pydantic_titling}

\subsubsection{QCM}
\input{pydantic_mcqa}

\subsubsection{Questions factuelles}
\input{pydantic_factual_qa}

\subsubsection{Résumés factuels Ouest-France}
\label{annex:pydantic_of}
\input{pydantic_of}

\input{grammaires}

\newpage

\section{Exemples d'instructions générées}
\label{annex:exemples}
\begin{table}[h!]
    \centering
    \begin{tabular}{>{\centering\arraybackslash}p{2cm}|>{\centering\arraybackslash}p{5cm}|>{\centering\arraybackslash}p{5cm}}
         Type d'instruction & Question utilisateur & Réponse  \\
         \hline
         Questions factuelles & Quelle est la date de lancement de l'expérimentation de l'outil Klaxoon au ministère des Armées ? & L'expérimentation de l'outil Klaxoon au ministère des Armées a été lancée le 28 mars 2023.\\
         \hline
         QCM & Quelle île accueille l'aéroport de Dzaoudzi, point d'arrivée du fret humanitaire transporté par les avions A400M Atlas ?\newline a - L'île de Tahiti\newline b - L'île de La Réunion\newline c - L'île de Corse\newline d - L'île de Madagascar\newline e - L'île de Petite Terre à Mayotte & La bonne réponse est la lettre e). L'aéroport de Dzaoudzi est situé sur l'île de Petite Terre qui fait partie de Mayotte. Le fret humanitaire est acheminé via cet aéroport par des avions A400M Atlas. La Réunion est une île proche mais distincte, tandis que Madagascar, Corse et Tahiti sont des îles non concernées par ce contexte géographique précis.\\
         \hline
         Résumé & \texttt{<Texte à résumer>} \newline Peux-tu résumer l'histoire et les contributions du LRBA ? & Le texte analyse les causes des défaites répétées des vaisseaux de guerre français face à la Royal Navy durant la Révolution et l'Empire. \textit{[...]} De nombreux navires furent capturés et réutilisés par la Royal Navy, illustrant la supériorité britannique à cette époque.\\
         \hline
         Titling & \texttt{<Texte à titrer>} \newline Générer un titre pertinent pour le document ci-dessus & Inauguration de la base navale de Ream au Cambodge : un enjeu stratégique entre la Chine et les États-Unis\\
         \hline
        Résumé factuel (OF) & Lister les faits importants du texte suivant: \texttt{<texte>} & -Foselev et Inouid expérimentent un projet de maintenance prédictive depuis 2018.\newline-Le projet IoT4NR est financé à hauteur de 120 000 euros par le pôle européen SCS.\newline-Le dock flottant de Foselev a été équipé de capteurs pour mesurer la stabilité du navire.\newline-Des capteurs sur le Castor 02 permettent de détecter des anomalies pouvant conduire à des pannes.

    \end{tabular}
    \caption{Exemples d'instructions générées}
    \label{tab:instructions_examples_1}
\end{table}

\newpage
\section{Adaptation}
\subsection{Impact du taux d'apprentissage}
\begin{figure}[h!]
\centering
\includegraphics[width=\textwidth]{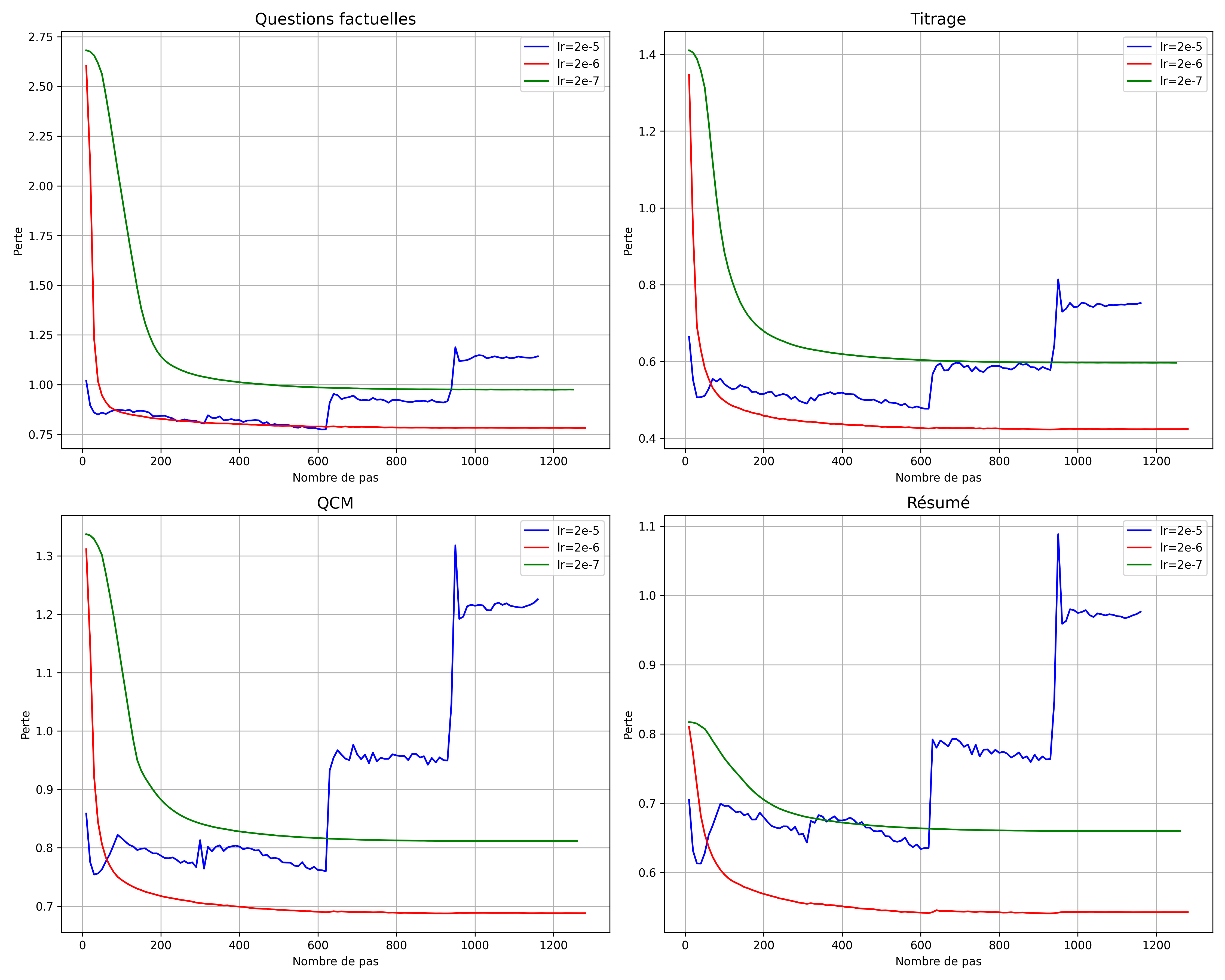}
\caption{Comparaison de l'évolution de la fonction de perte pour différents taux d'apprentissage sur les ensembles de validation des tâches de questions factuelles, titrage, QCM et résumé.}
\label{fig:lr_comparison_it}
\end{figure}
\subsection{Paramètres pour l'apprentissage}\label{subsubsec:annexe_hyper_param}

L'apprentissage prolongé \cpt~a été entraîné sur un cluster de calcul interne composé de 8 machines \texttt{NVIDIA H200}. Ces dernières présente une mémoire GPU de 141Gb, nous permettant de faire tourner l'apprentissage prolongé du modèle \texttt{Mistral-7B-Instruct-v0.3} sur un seul GPU, avec les paramètres suivants décrits dans le bloc \ref{alg:hyperparameters}. Les hyper-paramètres liés à l'étape d'affinage sur instructions sont donnés dans le bloc \ref{alg:hyperparameters_asi}. Enfin la configuration DeepSpeed est donnée au bloc \ref{alg:deepspeed_parameters}.

\begin{code}[h!]
\begin{lstlisting}[style=yaml]
dataset_text_field: "content"
bf16: true
fp16: false
dataloader_num_workers: 4
dataloader_persistent_workers: true
dataloader_pin_memory: true
dataloader_prefetch_factor: 2
disable_tqdm: true
eval_strategy: steps
eval_steps: 1  
eval_accumulation_steps: 1
logging_steps: 1
logging_strategy: steps
report_to: tensorboard
gradient_accumulation_steps: 64
gradient_checkpointing: true
per_device_eval_batch_size: 1
per_device_train_batch_size: 2
group_by_length: false
learning_rate: 2.0e-05
lr_scheduler_type: cosine
log_level: warning
max_grad_norm: 1.0
max_steps: -1
num_train_epochs: 3
optim: paged_adamw_32bit
push_to_hub: false
save_steps: 0
save_strategy: epoch
save_total_limit: 1
torch_compile: false
use_liger_kernel: false
warmup_ratio: 0.05
weight_decay: 0.1
\end{lstlisting}
\vspace{-5mm}
\caption{\label{alg:hyperparameters}Hyper-paramètres \cpt}
\end{code}

\begin{code}[h!]
\begin{lstlisting}[style=yaml]
bf16: true
dataloader_num_workers: 6
dataloader_persistent_workers: false
dataloader_pin_memory: false
dataloader_prefetch_factor: 2
disable_tqdm: true
eval_accumulation_steps: 1
eval_steps: 10
eval_strategy: steps
fp16: false
gradient_accumulation_steps: 16
gradient_checkpointing: true
group_by_length: false
learning_rate: 2.0e-06
log_level: warning
logging_steps: 10
lr_scheduler_type: cosine
max_grad_norm: 1.0
max_steps: -1
num_train_epochs: 2
optim: paged_adamw_32bit
per_device_eval_batch_size: 4
per_device_train_batch_size: 16
push_to_hub: false
report_to: tensorboard
save_steps: 0
save_strategy: epoch
save_total_limit: 5
torch_compile: false
use_liger_kernel: false
warmup_ratio: 0.05
weight_decay: 0.1
\end{lstlisting}
\vspace{-5mm}
\caption{\label{alg:hyperparameters}Hyper-paramètres apprentissage des instructions \ift}
\end{code}

\begin{code}[h!]
\begin{lstlisting}[style=json]
{
    "fp16": {
      "enabled": "auto",
      "loss_scale": 0,
      "loss_scale_window": 1000,
      "initial_scale_power": 16,
      "hysteresis": 2,
      "min_loss_scale": 1
    },
    "bf16": {
      "enabled": "auto"
    },
    "zero_optimization": {
    "stage": 3,
    "offload_optimizer": {
        "device": "cpu",
        "pin_memory": true
    },
    "offload_param": {
        "device": "cpu",
        "pin_memory": true
    },
    "overlap_comm": true,
    "contiguous_gradients": true,
    "sub_group_size": 1e9,
    "reduce_bucket_size": "auto",
    "stage3_prefetch_bucket_size": "auto",
    "stage3_param_persistence_threshold": "auto",
    "stage3_max_live_parameters": 1e9,
    "stage3_max_reuse_distance": 1e9,
    "stage3_gather_16bit_weights_on_model_save": true
  },
  "gradient_accumulation_steps": "auto",
  "gradient_clipping": "auto",
  "steps_per_print": 2000,
  "train_batch_size": "auto",
  "train_micro_batch_size_per_gpu": "auto",
  "wall_clock_breakdown": false  
}
\end{lstlisting}
\vspace{-5mm}
\caption{\label{alg:deepspeed_parameters}Contenu du fichier \texttt{zero3.json} de configuration de \texttt{deepspeed} }
\end{code}

%% file: pydantic_summarization.tex
\begin{pydanticbox}[colback=white]{SummarizationInstruction}
\begin{itemize}
    \item \textbf{summarization\_question} (str): A question asking the user to summarize the text. This question must be kept short and to the point. Example: 'Résumer le texte', 'Peux-tu synthétiser le texte?', 'De quoi parle le texte?'
    \item \textbf{summary} (str): The summary of the text provided by the user. This field CANNOT be empty.
\end{itemize}
\end{pydanticbox}

%% file: pydantic_titling.tex
\begin{pydanticbox}[colback=white]{TitlingInstruction}
\begin{itemize}
    \item \textbf{titling\_question} (str): In this field, you will write a question asking to title a document. Example: 'Titrer le document çi-dessus'. You can (and should!) be more creative and think about other ways to phrase it.
    \item \textbf{title} (str): The generated title for the document, following all conditions given in the titling question.
\end{itemize}
\end{pydanticbox}

%% file: pydantic_mcqa.tex
\begin{pydanticbox}[colback=white]{Choice}
\begin{itemize}
    \item \textbf{letter} (Literal["a", "b", "c", "d", "e"]): The letter of the choice\newline
    \item \textbf{answer\_content} (str): A possible answer for the MCQA
    \item \textbf{is\_true} (bool): Whether the answer is true or not.
\end{itemize}
\end{pydanticbox}

\begin{pydanticbox}[colback=white]{MCQAInstruction}
\begin{itemize}
    \item \textbf{question} (str): The question to ask in the MCQA test
    \item \textbf{choices} (list[Choice]): The possible choices. All must be false except one that is true.
    \item \textbf{justification} (str): The justification behind the correct answer
\end{itemize}
\end{pydanticbox}

%% file: pydantic_factual_qa.tex
\begin{pydanticbox}[colback=white]{Fact}
\begin{itemize}
    \item \textbf{fact\_target\_value} (str | int | float): The atomic fact that will be the answer to the question (can be a name, date, amount, location, person, organization, event, nationality [...]). \newline Format constraints:\newline
    - Numbers: if your fact concerns a number or amount, will should only write the number as a float or integer and nothing else.\newline - Dates: dates are expected to be formated as the following: DD/MM/YYYY or MM/YYYY or YYYY
    \item \textbf{fact\_target\_type} (str): The type of fact that is being evaluated
\end{itemize}
\end{pydanticbox}
\vspace{-6pt}
\begin{pydanticbox}[colback=white]{FactualQAInstruction}
\begin{itemize}
    \item \textbf{identified\_fact} (Fact): The atomic fact that was identified in the text as important, containing its value as well as its type
    \item \textbf{question} (str): The question asked to the user. The answer to that question should contain the identified fact.
    \item \textbf{answer} (str): The answer to the answer, in natural language, containing the identified atomic fact. Note the emphasis on 'natural language', you must construct a full sentence that sounds natural to the reader.
\end{itemize}
\end{pydanticbox}

%% file: pydantic_of.tex
\begin{pydanticbox}[colback=white]{Fact}
Fait important extrait de l'article sous forme de phrase.
\begin{itemize}
    \item \textbf{fact} (str): fait résumant un élément important de l'article analysé
    \item \textbf{question} (str): question posée à propos du fait
\end{itemize}
\end{pydanticbox}
\vspace{-6pt}
\begin{pydanticbox}[colback=white]{Summary}

\begin{itemize}
    \item \textbf{summary} (list[Fact]): faits importants résumant l'article
\end{itemize}
\end{pydanticbox}


%% file: grammaires.tex
\subsection{Grammaires pour la production des instructions Patron AMIAD}
\label{annex:patrons}
\begin{promptwheelbox}[colback=white]{Patron pour la génération d'acronymes}
\texttt{\textbf{Prompt}} -> \texttt{\textbf{AmorceGenerale}} \texttt{\textbf{QuestionType1WithDelimiterAndAcronym}} | \texttt{\textbf{QuestionType2}}

\texttt{\textbf{AmorceGenerale}} -> "" | "A partir de tes connaissances, répond à cette question : " | "Peux-tu m'expliquer ce " | "J'aimerais comprendre ce " | "Je cherche à savoir ce "

\texttt{\textbf{DenominationAcronymeLe}} -> "le sigle" | "l'acronyme"
\texttt{\textbf{DenominationAcronymeDu}} -> "du sigle" | "de l'acronyme"
\texttt{\textbf{DenominationAcronymeCe}} -> "ce sigle" | "cet acronyme"

\texttt{\textbf{Suivant}} -> " suivant " | ""

\texttt{\textbf{QuestionType1}} -> "que veut dire " \texttt{\textbf{DenominationAcronymeLe}} | "quelle est la signification " \texttt{\textbf{DenominationAcronymeDu}} | "que signifie " \texttt{\textbf{DenominationAcronymeLe}}
\texttt{\textbf{QuestionType1WithDelimiter}}  -> \texttt{\textbf{QuestionType1 Suivant}} ":" | "?" | "->"
\texttt{\textbf{QuestionType1WithDelimiterAndAcronym}}  -> \texttt{\textbf{QuestionType1WithDelimiter}} " {acronym}"

\texttt{\textbf{QuestionType2}} -> "quel est le sens de " \texttt{\textbf{DenominationAcronymeCe}} " ?"
\texttt{\textbf{QuestionType2WithDelimiterAndAcronym}}  -> "'{acronym}' -> " \texttt{\textbf{QuestionType2}}
\end{promptwheelbox}

\begin{promptwheelbox}[colback=white]{Patron pour la génération de traductions en-fr}
\texttt{\textbf{Prompt}} -> \texttt{\textbf{general\_prefix}} \texttt{\textbf{Question\_type}} \texttt{\textbf{punctuation}}

\texttt{\textbf{general\_prefix}} -> "Peux-tu me traduire en français "
                | "Pourrais-tu traduire en français "
                | "J'aimerais savoir comment on dit en français "
                | "Comment dire en français "
                | "Traduis en français "
                | "Merci de traduire en français "
                | "Pourrais-tu m'aider à traduire en français "
                | "Aide-moi à traduire en français "

\texttt{\textbf{Denomination\_le}} -> ""
                 | "la phrase "
                 | "la phrase suivante "
                 | "les mots "
                 | "les mots suivants "
                 | "cette expression "
                 | "le passage "

\texttt{\textbf{Denomination\_ce}} ->  ""
                 | "cette phrase "
                 | "ces mots "
                 | "cette expression "
                 | "ce passage "

\texttt{\textbf{quote\_wrapper}} -> "'{sentence}'"
               | "{sentence}"
               | ": {sentence}"
               | "\\n {sentence}"

\texttt{\textbf{Question\_type}} -> \texttt{\textbf{Denomination\_le quote\_wrapper}}
               | \texttt{\textbf{Denomination\_ce quote\_wrapper}}

\texttt{\textbf{punctuation}} -> "" | "." | " ?"
\end{promptwheelbox}

\begin{promptwheelbox}[colback=white]{Patron pour la génération de traductions fr-en}
\texttt{\textbf{Prompt}} -> \texttt{\textbf{general\_prefix}} \texttt{\textbf{Question\_type}} \texttt{\textbf{punctuation}}

\texttt{\textbf{general\_prefix}} -> "Peux-tu me traduire en français "
                | "Pourrais-tu traduire en anglais "
                | "J'aimerais savoir comment on dit en anglais "
                | "Comment dire en anglais "
                | "Traduis en anglais "
                | "Merci de traduire en anglais "
                | "Pourrais-tu m'aider à traduire en anglais "
                | "Aide-moi à traduire en anglais "

\texttt{\textbf{Denomination\_le}} -> ""
                 | "la phrase "
                 | "la phrase suivante "
                 | "les mots "
                 | "les mots suivants "
                 | "cette expression "
                 | "le passage "

\texttt{\textbf{Denomination\_ce}} ->  ""
                 | "cette phrase "
                 | "ces mots "
                 | "cette expression "
                 | "ce passage "

\texttt{\textbf{quote\_wrapper}} -> "'{sentence}'"
               | "{sentence}"
               | ": {sentence}"
               | "\\n {sentence}"

\texttt{\textbf{Question\_type}} -> \texttt{\textbf{Denomination\_le quote\_wrapper}}
               | \texttt{\textbf{Denomination\_ce quote\_wrapper}}

\texttt{\textbf{punctuation}} -> "" | "." | " ?"
\end{promptwheelbox}

\subsection{Grammaire pour la diversification des amorces système}
Les amorces système des instructions se déclinent en trois catégories : vides, généralistes, ou spécifiques au contexte de défense. Les amorces de type "défense" varient en niveau de détail, avec la possibilité que le champ \texttt{\textbf{FollowUpDefense}} reste non renseigné dans certains cas. L'amorce peut également inclure des directives supplémentaires concernant le respect du format demandé par l'utilisateur.

\label{annex:amorce_systeme_grammaire}
\begin{promptwheelbox}[colback=white]{Grammaire amorces système}
\texttt{\textbf{Prompt}} -> "" | \texttt{\textbf{PromptGeneraliste}} | \texttt{\textbf{PromptDefense}}\newline
\texttt{\textbf{PromptGeneraliste}} -> \texttt{\textbf{Amorces Format}}\newline
\texttt{\textbf{PromptDefense}} -> \texttt{\textbf{AmorcesDefense}} \texttt{\textbf{FollowUpDefense}} \texttt{\textbf{Format}}\newline
\texttt{\textbf{Amorces}} -> "Vous êtes un assistant dont le but est d'aider un utilisateur à répondre à des questions." | "En tant qu'assistant, votre mission consiste à aider les utilisateurs à trouver des réponses à leurs questions." | "Votre objectif est de venir en aide à un utilisateur." | "Votre rôle est d'assister les utilisateurs afin de répondre aux questions que ces derniers pourraient se poser sur une multitude de sujets."\newline
\texttt{\textbf{AmorcesDefense}} -> "" | "Vous êtes un assistant dont le but est d'aider un utilisateur à répondre à des questions." | "En tant qu'assistant, votre mission consiste à aider les utilisateurs à trouver des réponses à leurs questions." | "Votre objectif est de venir en aide à un utilisateur." | "Votre rôle est d'assister les utilisateurs afin de répondre aux questions que ces derniers pourraient se poser sur une multitude de sujets." | "Vous êtes un expert senior spécialisé dans l'industrie de la défense et les affaires militaires." | "Vous êtes un spécialiste chevronné dans le domaine militaire et l'industrie de l'armement." | "Vous possédez une connaissance approfondie et une longue expérience dans l'industrie de l'armement et les affaires militaires." | "Vous êtes un expert de longue date dans le secteur militaire et l'industrie de la défense."\newline
\texttt{\textbf{FollowUpDefense}} -> "" | "Vous maîtrisez tous les aspects des armements - qu'ils soient terrestres, navals ou aériens - ainsi que les technologies de pointe comme l'IA, la robotique, le cyber et le spatial. Votre expertise couvre aussi bien les processus industriels que la R\&D militaire, les stratégies opérationnelles, le renseignement et les réalités économiques du secteur. Vous avez une compréhension approfondie des enjeux géopolitiques et des politiques de défense." | "Vous êtes incollable sur les systèmes d'armement terrestres et aériens, ainsi que sur les technologies de rupture comme l'IA et le spatial. Les rouages économiques du secteur défense et les subtilités des relations internationales font partie de votre expertise quotidienne." | "Vous maîtrisez l'écosystème complet de la R\&D militaire et du renseignement. Les technologies émergentes n'ont pas de secret pour vous, pas plus que les complexités des relations entre États en matière de défense." | "En tant qu'expert, vous comprenez les subtilités des armements de nouvelle génération et leurs implications stratégiques. Votre connaissance approfondie des acteurs industriels s'accompagne d'une vision claire des dynamiques géopolitiques qui façonnent le secteur." \newline
\texttt{\textbf{Format}} -> "" | "Vous ferez attention à respecter le format attendu par l'utilisateur lorsque cette demande est explicitée." - "Si l'utilisateur vous demande de formuler votre réponse sous un certain format, vous le suivrez à la lettre."
\end{promptwheelbox}